\documentclass[11pt]{article}

\usepackage[preprint]{acl}

\usepackage{times}
\usepackage{latexsym}
\usepackage{booktabs}
\usepackage{makecell}
\usepackage{tabularx}
\usepackage{multirow}
\usepackage{enumitem}
\usepackage{color} 
\usepackage{soul}
\usepackage{array}
\usepackage{tikz}
\newcommand*\circled[1]{\tikz[baseline=(char.base)]{
            \node[shape=circle,draw,inner sep=1pt] (char) {#1};}}
\usepackage[T1]{fontenc}

\usepackage[utf8]{inputenc}

\usepackage{microtype}

\usepackage{inconsolata}

\usepackage{graphicx}

\usepackage{tcolorbox}
\tcbuselibrary{skins}
\usepackage{dblfloatfix}
\usepackage{makecell}
\usepackage{dblfloatfix}

\definecolor{CB_lightCyan}{HTML}{99DDFF}
\definecolor{CB_darkCyan}{HTML}{66CCFF}
\definecolor{CB_pear}{HTML}{BBCC33}
\definecolor{CB_pink}{HTML}{E4CEE1}

\newcommand{\ip}[1]{\textcolor{purple}{}}

\title{Explain Like I'm 5 or Whatever I Choose: Evaluating the Potential for Interactive Complexity in Language Model Responses}

\author{Indu Panigrahi and Tal August \\
  Siebel School of Computing and Data Science \\
  University of Illinois Urbana-Champaign\\
  \texttt{\{indup2, taugust\}@illinois.edu} }

\begin{document}

\tcbset{taggingPrompt/.style={
    enhanced,
    size=fbox,
    boxrule=2pt,
    arc=2mm,
    auto outer arc,
    left=1pt,
    right=1pt,
    top=1pt,
    bottom=1pt,
    fontupper=\ttfamily,
    colback=CB_pink!15,
    colframe=CB_pink!30,
    coltitle=CB_pink!25!black,
}}


\maketitle


\begin{abstract}
    Given the increased use of large language models (LLMs) for reading and seeking information and the varying needs and backgrounds of users, direct manipulation of text can be a valuable mechanism for end-user control.
    However, enabling direct manipulation requires that models generate text consistent with user-requested changes.
    We focus on one such setting, manipulating the complexity of scientific information, and ask if models can generate multiple summaries along a continuum of text complexity that enable effective user selection and control.
    Through a within-subjects user study, we identify reliable complexity measures and a criterion for how models should conform to these measures.
    We evaluate five LLMs of different families, sizes, and reasoning abilities on $557$ scientific queries. We find that while models can vary complexity across responses, most changes remain inconsistent, with models shifting the complexity measures in the correct direction at most $33.67\%$ of the time.

\end{abstract}


\section{Introduction}

Scientists and students are increasingly using LLMs for reading \citep{Fok2023scim, Lo2023TheSR} and synthesizing literature \citep{asai2024openscholar, ChatGPTDeepResearch}. 
Readers vary in their preferred responses based on their background (e.g., a junior vs. senior researcher) and context (e.g., reading within a familiar topic vs. new discipline), preferring simpler or more complex summaries and explanations \citep{guo2021biomedsumm, august2022generating, joshi2025eliwhy}.\looseness=-1

Past evaluations have focused on models' ability to generate responses that align with different envisioned audiences \citep{joshi2025eliwhy}, or personalize a response to a given user \citep{guo-etal-2024-personalized, murthy-etal-2021-personalized}. However, a single response can be misaligned with a user or fail to incorporate correct context \citep{Gu2024airesilient}. Further, responses may be inherently insufficient even if correct \citep{fok2023qlarify}. For example, the design paradigm of \textit{details on demand} \citep{min2025malleable} suggests that users may---after seeing an initial summary---want more details, even if they did not want a detailed first response. While users might request more details, repetitive prompting is time-intensive \citep{10.1145/3544548.3581388}. Direct manipulation of text (e.g., a slider for text complexity) can be a more effective mechanism for end-user control \citep{zhang2026wordswidgetscontrollablellm}.\looseness=-1

From an evaluation standpoint, this requires a shift towards evaluating the quality of \textit{multiple} responses and how they relate to each other, rather than of single responses \cite{joshi2025eliwhy, murthy-etal-2021-personalized}.
Thus, rather than asking if models can generate a single best scientific summary, in this paper we ask if models can generate \textit{multiple} summaries that enable effective user selection and control.
Triangulating with prior work, we identify measures of language complexity through a within-subjects user study where participants ($N=16$) explored unfamiliar STEM\footnote{Science, Technology, Engineering, and Mathematics} topics using a prototype chat interface that enabled direct manipulation of response language complexity (Fig. \ref{fig:cui}).
Using an existing scientific QA benchmark \citep{asai2024openscholar}, we test $5$ models of different sizes, families, and reasoning capabilities (GPT-5.1, GPT-5 mini, Claude Sonnet 4.5 $\pm$ Thinking, and DeepSeek-V3.1) by generating $5$ versions, at different levels of response complexity\footnote{We label complexity levels in our prompt by using different envisioned audiences, similar to past work \citep{joshi2025eliwhy, august2022generating}, see Sec \ref{sec:userstudy_method}.}, of answers to $557$ scientific queries.
We define model performance based on the relationship \textit{between} versions of a response.\looseness=-1

We find that while models are able to vary complexity across responses, most changes remain inconsistent.
For example, one of the better-performing models, Claude Sonnet 4.5 + Thinking, increases complexity $17.35\%$ of the time across the five versions for an individual query (and at most $76.53\%$ of the time for a single transition between versions).
Additionally, models increase complexity measures less consistently between higher complexity responses than between lower complexity responses (e.g., $80.61\%$ of the time between the first two versions reduces to $43.88\%$ between the last two versions for Claude Sonnet 4.5).
In summary, our paper makes the following contributions:\looseness=-1

\begin{enumerate}
    \item A new evaluation approach that tests models' ability to generate multiple versions of a response across a dimension of interest. We instantiate the framework for scientific information seeking and language complexity.

    \item A suite of three complexity measures motivated from prior work and grounded to a user study with $16$ participants reading scientific literature in unfamiliar disciplines. To evaluate models, we define a criterion for the relationship of complexity measures \textit{between} versions of a response, further justified with additional human validation.

    \item Evaluation results from $5$ models showing that models often fail to reliably adjust complexity measures in the correct direction across versions. Our findings hold across models and when we shift the complexity levels.
\end{enumerate}

\section{Related Work}

\subsection{LLMs for Information-Seeking}
With the rising capabilities of LLMs being able to quickly produce different versions of text~\citep{Kirk2024alignment,August2024know,joshi2025eliwhy,wu2023summarization}, many LLM-powered interfaces have been developed to help people search for papers~\citep{mudd2025screen}, skim papers~\citep{Fok2023scim}, aggregate information across multiple documents~\citep{singh2025scholarqa,Whitfield2023elicit}, and understand content~\citep{August2023paperplain,Rust2025llmsummary,Fok2023scim,rust2025lifestyle,fok2023qlarify}.
A popular option has been conversation-based, question-answering systems, such as Elicit~\citep{Whitfield2023elicit}, ASTA~\citep{singh2025scholarqa}, and OpenAI's Deep Research~\citep{ChatGPTDeepResearch}.\looseness=-1

\subsection{Adapting LLM Responses to Users}

There have been two primary ways to adapt LLM responses to users: interactivity (i.e., the \textit{user decides} what to see)~\citep{min2025malleable,Sundar2010interactivity,farber2025simplifymytext,head2021augment} and personalization (i.e., the \textit{system decides} what the user sees)~\citep{kim2025translider,adar2017persalog,august2022generating,acharya2018towards}. Past evaluations in complexity adaption have had an implicit \textit{personalization} context, meaning the goal was for the model to generate a single correct response, evaluating whether or not LLMs could simplify text by generating a response at a 5th grade reading level when prompted with ``\textit{explain like I am a 5th grader}'' for example~\citep{beks2024govt,Hedlin2025learning,joshi2025eliwhy,farber2025simplifymytext}.
We instead focus on an \textit{interactive} context by evaluating how well models generate a range of responses, allowing users to choose between levels of complexity. We explore this alternative framing because users may need to select their own version depending on their needs in the moment \citep{fok2023qlarify} rather than on a static attribute (e.g., a professor wanting a simple definition of a term in their field). While users can prompt models for this information, specifying information needs iteratively can be time-consuming and distracting \citep{10.1145/3544548.3581388}.\looseness=-1

\section{Formative User Study}

Inspired by small-scale formative studies in Human-Computer Interaction research \citep{kim2023hmhta,Sauro_Lewis_2012}, and their growing use in NLP to identify a contextualized task of interest to then evaluate at scale \citep{guo-etal-2024-personalized,kim2023hmhta, bhatt-etal-2025-research}, we conduct a formative study with $16$ participants to structure our subsequent model evaluation (Sec. \ref{sec:model_eval_method}).
Our formative study: (1) validates the utility of interactive complexity in the context of scientific information seeking, (2) corroborates characteristics of language associated with user-perceived complexity, and (3) identifies requirements for model responses that enable successful interactive complexity.\looseness=-1


\subsection{Study Design and Analysis}
\label{sec:userstudy_method}

\begin{figure}[h]
    \centering
    \includegraphics[width=\linewidth]{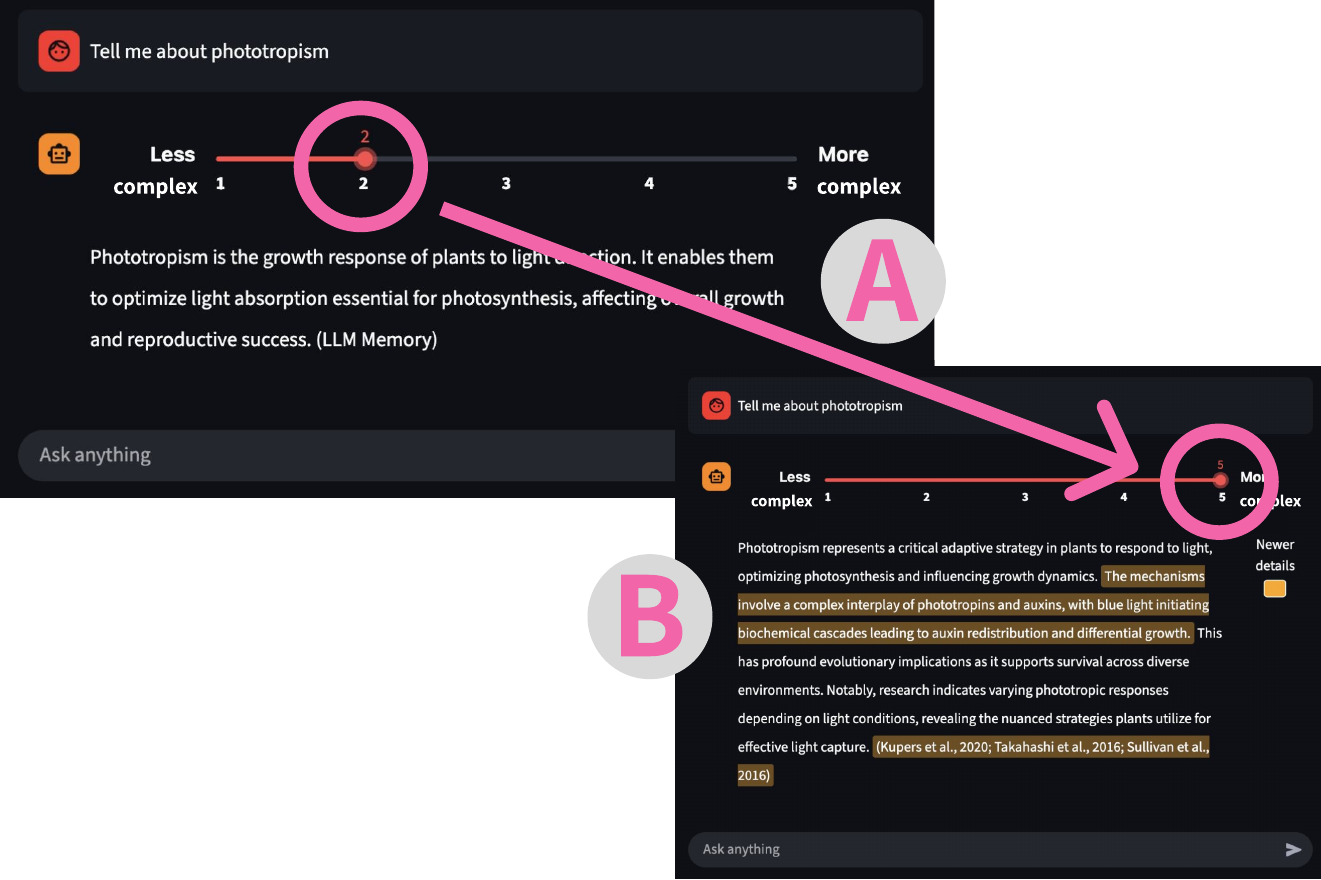}
    \caption{\textbf{Interactive language complexity interface} Users can manipulate textual complexity by moving the response slider (A) to a different notch. Sentences that are significantly different from those in the previously-displayed version are highlighted (B); significant differences are determined by comparing sentence-level BERTScores \citep{bertscore} to a threshold. The preset default is 3, and there is an option to change the default to apply to all sliders.}
    \label{fig:cui}
\end{figure}

\paragraph{Procedure} We recruited $16$ participants to interact with two interfaces. An interactive complexity chat provided users with a slider mechanism to adjust the language complexity of responses, choosing from $5$ levels, $1$ being the least complex (Fig. \ref{fig:cui}).
The conventional interface looked and functioned similar to a basic chat interface (i.e., without the slider). All studies were counterbalanced between the two interfaces.

To emulate the experience of exploring new, knowledge-intensive topics with the interfaces, we asked each participant to provide two STEM topics before the study that they had interest in but little to no knowledge about. For each interface, participants were instructed to identify facts and a paper within their chosen topics.\looseness=-1

While interacting with each interface, participants actively verbalized their reasoning and impressions.
After each condition, we interviewed participants about their experience with the interface.
More details and materials are provided in Appendix \ref{sec:user_study_materials_app}.\looseness=-1

\paragraph{Response Generation} We used the participant-provided topics to generate scientific reports using a RAG-based pipeline \citep[ScholarQA,][]{singh2025scholarqa} with Claude Sonnet 4.5. Using these reports as inputs, we prompted GPT-5 mini\footnote{Chosen for low latency appropriate for an interactive context.} to generate $5$ versions of the reports for audiences defined as College student, Junior Ph.D. student, Senior Ph.D. student, Postdoctoral researcher, and Senior researcher\footnote{We based these levels off of previous work that stratifies model responses using level of education \citep{joshi2025eliwhy,August2024know}, adjusting the levels to be appropriate for the envisioned prompts (i.e., scientific literature queries).} to populate the slider. Appendix \ref{sec:prompt} includes the prompt and describes alternative prompts we tested.\looseness=-1

\paragraph{Analysis} To analyze the participant responses, we employ open coding \citep{Saldana2021coding} (details in Appendix \ref{sec:qual_coding_app}).
Since all authors were on consultation for study coding, we did not calculate inter-rater reliability \citep{McDonald2019irr}.
The codes describe participants' sentiments (e.g., ``\textit{Appreciates agency of interactive chat}'') and interactions (e.g., ``\textit{Prompted with desired complexity}'').\looseness=-1





\subsection{Results}
\label{sec:user_study_results}

\paragraph{Interactive complexity is valuable over a conventional chat interface}
The majority of participants (13/16) appreciated the flexibility that interactive complexity provided beyond prompting. Participants found the responses from the conventional chat interface inconsistent in terms of complexity, sometimes providing too much or too little information or jargon (13/16). In fact, seven participants (four of whom had not yet seen the interactive condition) ended up describing their preferred level of complexity in follow-up prompts, indicating a strong desire for control over perceived complexity of responses.

\paragraph{Jargon, information, and length influenced perceived complexity}
As participants decreased complexity, they prioritized three primary, desired trends: decreases in scientific jargon (16/16), amount of information (13/16), and length (12/16).
This finding reinforces what prior work has proposed for measuring complexity \citep{August2024know,joshi2025eliwhy,appls} and forefronts the measures that we focus on in our model evaluation (Sec. \ref{sec:model_eval_method}).\looseness=-1

\paragraph{Small changes in complexity are hard to perceive}
While five levels of complexity seemed generally appropriate (10/16), ten participants noted the levels needed to be more distinct from each other. 
In particular, six participants shared the sentiment that P9 expressed of ``\textit{one and two [the least complex versions] and four and five [didn't] feel all that much different}''.
This observation suggests that some responses did not strictly increase in complexity, conflicting with what users expect and need.\looseness=-1

\section{Evaluating Interactive Complexity}
\label{sec:model_eval_method}
Motivated by our formative findings, we test the potential for models to provide responses that enable interactive complexity. Below we describe the data (Sec. \ref{sec:data}) and models (Sec. \ref{sec:models}) used in our evaluation, as well as our measures of complexity (Sec. \ref{sec:measures}) and our criterion to facilitate effective user control over model responses (Sec. \ref{sec:criterion}).\looseness=-1

\subsection{Scientific Query Data}
\label{sec:data}
We use $557$ queries from ScholarQABench~\citep{asai2024openscholar} which contains expert-written scientific queries across multiple STEM fields (Appendix \ref{sec:scholarqabench_app}).
$98$ of these queries come with expert-written answer reports.
We restrict one evaluation to this subset to use the expert-written reports for grounding model responses to a single ground truth report (similar to the user study); this choice aligns with past work on the utility of human-written context \citep{tan2024humancontext,zhang2023ragcontext} and potential issues with reusing models for both initial ground truth generation and subsequent response generation \citep{xu2024llmselfbias,2024biased}.
Since this restriction leads to a relatively small sample, we run a second evaluation on a larger sample of queries from the rest of ScholarQABench with reports generated from ScholarQA \citep{singh2025scholarqa}, similar to the user study.
Specifically, we randomly sample $500$ scientific queries from ScholarQABench (excluding the aforementioned $98$) and generate a report for each using the ScholarQA pipeline \citep{singh2025scholarqa} with a Claude Sonnet 4.5 backend. 
We use these reports as inputs to each model.
We report results on $459$ queries due to model refusal on some of the queries.\looseness=-1


\subsection{Models}
\label{sec:models}
We prompt models the same way as for the user study (Sec.~\ref{sec:userstudy_method})---to generate $5$ versions of the input, stratified for five envisioned audiences going from least to most complex (College student $\rightarrow$ Senior researcher). Since we label each version with an intended audience, we run a third evaluation that uses alternative audience labels (Child $\rightarrow$ Expert) (Sec.~\ref{sec:results_ext}). We evaluate $5$ models across different sizes, model families, and reasoning abilities: GPT-5.1, GPT-5 mini, Claude Sonnet 4.5 (with and without Thinking), and DeepSeek-V3.1. Details about model configurations are in Appendix \ref{sec:model_config}.\looseness=-1

\subsection{Complexity Measures}
\label{sec:measures}
Past work has quantified complexity through several linguistic measures. These include readability formulas \cite{flesch1948}, lexical features, and perplexity measures \citep{appls, August2024know, joshi2025eliwhy}. We triangulate between these works and our formative study results (Sec.~\ref{sec:user_study_results}) to curate a suite of three measures,\footnote{The Flesch-Kincaid Reading Ease Score~\cite{flesch1948} is a commonly-used scale for rating complexity; however, past work has shown that Flesch-Kincaid scores do not provide a reliable measure of complexity~\cite{cachola2025flesch}. We focus on the measures confirmed in our user study and provide Flesch-Kincaid data (reflects similar trends as the other measures) in Appendix \ref{sec:flesch_data}.} each representing different dimensions of perceived language complexity:\looseness=-1

\begin{itemize}
    \item \textsc{Jargon} \cite{appls,joshi2025eliwhy}: This measure denotes the number of words that are considered specialized or technical. We calculate the percentage of words in the response that are not on the Dale-Chall Word List, a list of $3{,}000$ familiar English words~\citep{Chall1995ReadabilityR}.\looseness=-1
    \item \textsc{Information} \cite{trienes2024infolossqa,appls}: More complex language often includes more information (e.g., technical details). We query GPT-4.1 to identify the number of independent facts, using the pipeline for generating ``atomic facts'' from \citet{min2023factscore}; we validate the performance by manually inspecting a subset of $25$ examples (Appendix \ref{sec:info_validation_app}).\looseness=-1
    \item \textsc{Length} \citep{joshi2025eliwhy,appls}: More complex language, similar to containing more information, is often longer. However, simpler language can also be longer when the same information is explained or elaborated\footnote{We consider information density (the ratio between \textsc{Information} and \textsc{Length}) as way to measure elaborative simplification. However, we already include both measures separately, and model performance was lower than and reflected the same trends as with our selected measures (Appendix \ref{sec:density_app})} on \citep{wu2023elaborative}. We quantify length as the total number of response characters.
\end{itemize}

\subsection{Criterion for effective user control}
\label{sec:criterion}

Based on prior work~\citep{appls,joshi2025eliwhy,August2024know} and our formative user study, we define model performance as a model's ability to increase complexity measures as the intended complexity levels increase.
We operationalize this by evaluating the \textit{direction of changes} in the complexity measures across the $5$ versions of a response.
\textbf{Models that are better at generating distinct levels of complexity will produce positive changes in all measures.}

Our criterion rests on the assumption that our measures of complexity capture a user's perception of complexity. To validate this assumption, we perform a comparison between the complexity measures and human ratings of perceived complexity.
We randomly sampled $100$ pairs of generated text from consecutive levels in our model evaluation (evenly sampled across the four transitions and five models) and conducted manual binary comparisons of which text was perceived as more complex. Comparisons were done by two authors in the research team in a blind setting\footnote{The texts within each sampled pair were shuffled, and two authors rated the first $25$ pairs. Since the authors achieved high agreement ($88\%$, Cohen’s $\kappa$ = $0.75$), one author rated the remaining $75$ pairs.}.
Importantly, ratings were not focused on measuring \textsc{Jargon}, \textsc{Information}, or \textsc{Length}, but rather a user's overall perceived complexity.
We compare agreement, measured as Cohen's $\kappa$\footnote{Chosen for two raters classifying samples according to a binary category.}, between these human ratings and the complexity measures (Tab. \ref{tab:criterion_agreement}).
We find that the agreement between the ratings and the combination of all measures is substantial ($\kappa = 0.84$), and much higher than any single measure alone (e.g., \textsc{Jargon} alone is $\kappa = 0.56$).
This further emphasizes the importance of all measures increasing together as opposed to individually.\looseness=-1

\renewcommand{\arraystretch}{1}
\begin{table}[h]
    \centering
    \small
    \begin{tabular}{l|c|c}
        \textbf{Measure(s)} & \textbf{Agreement} & \textbf{\# Agreed / \# Inputs} \\
        \toprule
        Jargon & 0.56 & 78 / 100 \\
        + Information & 0.77 & 53 / 60 \\
        + Length & \underline{0.83} & 64 / 70 \\
        \midrule
        Information & 0.36 & 68 / 100 \\
        + Length & 0.69 & 59 / 70 \\
        \midrule
        Length & 0.60 & 80 / 100 \\
        \midrule
        All & \textbf{0.84} & 46 / 50 \\
        \bottomrule
    \end{tabular}
    \caption{\textbf{Agreement on measures for criterion} Each row corresponds to either a single measure (e.g., \textsc{Jargon}) or a combination of measures (e.g., \textsc{Jargon} + \textsc{Information}). ``Agreement'' indicates the agreement between the measure(s) and the human rater as Cohen's $\kappa$. ``\# Agreed / \# Inputs'' indicates how many inputs the ratings and measure(s) agree on out of the number of inputs that the measure(s) agree on.}
    \label{tab:criterion_agreement}
\end{table}

\section{Results}
\label{sec:model_eval_results}
%

\begin{figure*}[t]
    \centering
    \includegraphics[width=\linewidth]{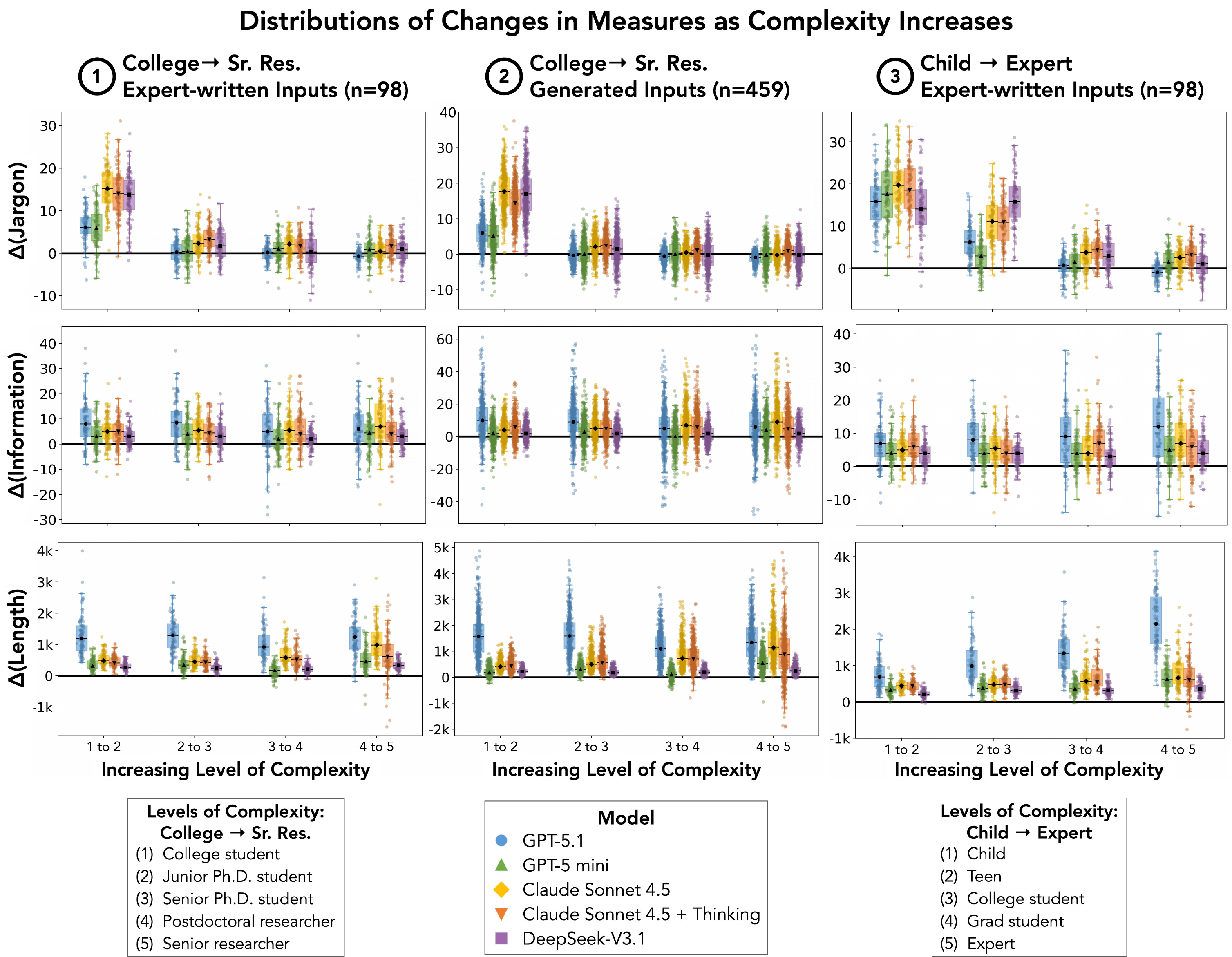}
    \caption{\textbf{Model performance shown as changes in complexity measures for the three evaluations} Each point in the scatter overlays represents an input. Between consecutive levels of complexity, models produce changes in \textsc{Jargon}, \textsc{Information}, and \textsc{Length} that vary between increasing and decreasing, particularly for \textsc{Jargon} and \textsc{Information}, regardless of input source and audience labels. Extreme outliers removed for visualization.
    }
    \label{fig:boxplots}
\end{figure*}

We present results from the three evaluations, varying the input source (expert-written and model-generated) and the audience labels (College student $\rightarrow$ Senior researcher and Child $\rightarrow$ Expert).
Since the two evaluations that vary the input source share the same trends, we discuss them together throughout Sec. \ref{sec:result_inconsistent}-\ref{sec:result_later}.
Sec. \ref{sec:results_ext} discusses the evaluation of alternate audience labels.\looseness=-1

Fig.~\ref{fig:boxplots} plots the change in each measure between consecutive levels of complexity (descriptive statistics provided in Appendix \ref{sec:descriptive_stats_app}).
Tab.~\ref{tab:model_performance_percentages} reports the percent of changes that move in the correct direction (i.e., increasing \textsc{Jargon}, \textsc{Information}, or \textsc{Length}).
Because Fig. \ref{fig:boxplots} and Tab.~\ref{tab:model_performance_percentages} separate results by the four transitions between levels, we also report the overall performance across all transitions in Tab.~\ref{tab:model_performance_good_combined} (i.e., the percent of inputs for which each model moves in the correct direction at all transitions between the five levels). 
To test if the distribution of changes moving in the wrong direction is significantly different from the distribution of changes moving in the right direction at each transition, we run Welch's $t$-tests, with a Bonferroni correction, comparing those distributions at each of the four transitions (Appendix \ref{sec:sigtest_app}).
Lastly, we confirm in Appendix \ref{sec:user_study_complexity} that the distribution of the responses we report here reflects the response distribution from our user study.
We also check factual consistency and consistency between runs (Appendix \ref{sec:validity_app}). \looseness=-1

\renewcommand{\arraystretch}{1}
\setlength{\tabcolsep}{4pt}
\begin{table*}[bt]
    \centering
    \small
    \begin{tabular}{cl!{\vrule width 2pt}c|c|c|c|}
        \multicolumn{2}{l}{} & \multicolumn{4}{l}{\textbf{\circled{1} \makecell[l]{College $\rightarrow$ Sr. Res.\\Expert-written (n=98)}}} \\
    \midrule
        & \textbf{Model} & \textbf{Jar.} & \textbf{Inf.} & \textbf{Len.}  & \textbf{All} \\ 
    \midrule
         & GPT-5.1 & 94.90 & \textbf{81.63} & \textbf{100.0} & 77.55 \\
         \multirow{1}{*}[-5pt]{\rotatebox{90}{\textbf{1 to 2}}} &GPT-5 mini & 87.76 & 72.45 & \textbf{100.0} & 62.25 \\
         &Sonnet 4.5 & \textbf{100.0} & 80.61 & \textbf{100.0} & \textbf{80.61} \\
         &\makecell{ + Thinking} & 98.98 & 77.55 & \textbf{100.0} & 76.53 \\
         &DeepSeek-V3.1 & 98.98 & 77.55 & \textbf{100.0} & 76.53 \\

    \midrule
         & GPT-5.1 & 54.08 & \textbf{83.67} & \textbf{100.0} & 43.88 \\
         \multirow{1}{*}[-5pt]{\rotatebox{90}{\textbf{2 to 3}}} &GPT-5 mini & 57.14 & 74.49 & 97.96 & 42.86 \\
         &Sonnet 4.5 & 83.67 & 80.61 & \textbf{100.0} & 67.35 \\
         &\makecell{ + Thinking} & \textbf{88.78} & 79.59 & \textbf{100.0} & \textbf{71.43} \\
         &DeepSeek-V3.1 & 69.39 & 74.49 & \textbf{100.0} & 54.08 \\

    \midrule
         & GPT-5.1 & 55.10 & 71.43 & \textbf{100.0} & 38.78 \\
         \multirow{1}{*}[-5pt]{\rotatebox{90}{\textbf{3 to 4}}} &GPT-5 mini & 65.31 & 57.14 & 80.61 & 27.55 \\
         &Sonnet 4.5 & 71.42 & \textbf{78.57} & \textbf{100.0} & 55.10 \\
         &\makecell{ + Thinking} & \textbf{77.55} & 75.51 & 98.98 & \textbf{58.16} \\
         &DeepSeek-V3.1 & 51.02 & 69.39 & 98.98 & 35.71 \\

    \midrule
         & GPT-5.1 & 27.55 & \textbf{77.55} & 97.96 & 18.37 \\
         \multirow{1}{*}[-5pt]{\rotatebox{90}{\textbf{4 to 5}}} &GPT-5 mini & 60.20 & 69.39 & 95.92 & 41.84 \\
         &Sonnet 4.5 & 55.10 & 83.67 & \textbf{100.0} & 43.88 \\
         &\makecell{ + Thinking} & \textbf{76.53} & 76.53 & 86.73 & \textbf{55.10} \\
         &DeepSeek-V3.1 & 60.20 & 75.51 & \textbf{100.0} & 47.96 \\

    \bottomrule
    \end{tabular}
    \hspace{0.2em}
    \begin{tabular}{!{\vrule width 2pt}c|c|c|c|}
        \multicolumn{4}{l}{\textbf{\circled{2} \makecell[l]{College $\rightarrow$ Sr. Res.\\Generated (n=459)}}} \\
    \midrule
        \textbf{Jar.} & \textbf{Inf.} & \textbf{Len.} & \textbf{All} \\
    \midrule
         93.03 & \textbf{84.75} & \textbf{100.0} & 79.09 \\
         86.06 & 59.48 & 92.37 & 48.58 \\
         \textbf{100.0} & 76.03 & \textbf{100.0} & 76.04 \\
         \textbf{100.0} & 81.05 & 99.78 & \textbf{81.05} \\
         99.56 & 69.50 & 99.56 & 69.06 \\
    \midrule
         44.01 & \textbf{79.30} & \textbf{100.0} & 35.29 \\
         51.20 & 66.01 & 99.35 & 33.33 \\
         74.29 & 76.69 & \textbf{100.0} & 56.43 \\
         \textbf{79.74} & 76.47 & 99.78 & \textbf{62.75} \\
         62.96 & 61.22 & \textbf{100.0} & 38.78 \\
    \midrule
         35.51 & 67.76 & \textbf{100.0} & 25.49 \\
         50.76 & 48.58 & 77.34 & 20.26 \\
         56.43 & \textbf{80.17} & \textbf{100.0} & 45.10 \\
         \textbf{63.83} & 76.03 & 98.04 & \textbf{48.58} \\
         48.37 & 59.04 & 99.35 & 28.32 \\
    \midrule
         25.05 & 70.59 & 99.56 & 17.43 \\
         47.93 & 69.28 & 98.04 & 32.24 \\
         45.32 & \textbf{79.96} & \textbf{100.0} & 34.86 \\
         \textbf{67.32} & 66.45 & 85.19 & \textbf{38.56} \\
         45.97 & 64.92 & 99.56 & 29.85 \\
    \bottomrule
    \end{tabular}
    \hspace{0.2em}
    \begin{tabular}{!{\vrule width 2pt}c|c|c|c|}
        \multicolumn{4}{l}{\textbf{\circled{3} \makecell[l]{Child $\rightarrow$ Expert\\Expert-written (n=98)}}} \\
    \midrule
        \textbf{Jar.} & \textbf{Inf.} & \textbf{Len.} & \textbf{All} \\
    \midrule
         \textbf{100.0} & 81.63 & \textbf{100.0} & 81.63 \\
         98.98 & 81.63 & 98.98 & 80.61 \\
         \textbf{100.0} & 87.76 & \textbf{100.0} & 87.76 \\
         \textbf{100.0} & \textbf{91.84} & \textbf{100.0} & \textbf{91.84} \\
         97.96 & 87.76 & 97.96 & 83.67 \\

    \midrule
         95.92 & \textbf{92.86} & \textbf{100.0} & \textbf{88.78} \\
         75.51 & 80.61 & \textbf{100.0} & 60.20 \\
         97.96 & 80.61 & \textbf{100.0} & 79.59 \\
         97.96 & 81.63 & \textbf{100.0} & 79.59 \\
         \textbf{100.0} & 81.63 & \textbf{100.0} & 81.63 \\

    \midrule
         63.27 & 78.57 & \textbf{100.0} & 50.00 \\
         65.31 & 73.47 & 98.98 & 51.02 \\
         79.59 & 83.67 & \textbf{100.0} & 66.33 \\
         \textbf{88.78} & \textbf{84.69} & \textbf{100.0} & \textbf{75.51} \\
         78.57 & 73.47 & \textbf{100.0} & 57.14 \\

    \midrule
         36.73 & \textbf{85.71} & \textbf{100.0} & 32.65 \\
         69.39 & 81.63 & 97.96 & 56.12 \\
         78.57 & 83.67 & \textbf{100.0} & \textbf{65.31} \\
         \textbf{92.86} & 71.43 & 93.88 & \textbf{65.31} \\
         64.29 & 82.65 & \textbf{100.0} & 52.04 \\

    \bottomrule
    \end{tabular}

    \caption{\textbf{Model performance per transition} Each model's performance is shown as the percent of inputs for which the measure moves in the correct direction at each transition; ``All'' means where all three measures move in the correct direction. A higher percentage means that the model performed better by more often increasing complexity. 
    }
    \label{tab:model_performance_percentages}
\end{table*}

\renewcommand{\arraystretch}{1}
\setlength{\tabcolsep}{6pt}
\begin{table}[t]
    \centering
    \small
    \begin{tabular}{l|c|c|c|c}
    \multicolumn{5}{l}{\textbf{\circled{1} College $\rightarrow$ Sr. Res., Expert-written (n=98)}} \\
    \midrule
    \textbf{Model} & \textbf{Jargon} & \textbf{Info.} & \textbf{Length} & \textbf{All} \\
    \midrule
         GPT-5.1 & 5.10 & 31.63 & 97.96 & 3.06 \\
         GPT-5 mini & 13.27 & 14.29 & 76.53 & 0.00 \\
        Sonnet 4.5 & 32.65 & \textbf{42.86} & \textbf{100.0} & 14.29 \\
        \makecell{ + Thinking} & \textbf{45.92} & 32.65 & 86.73 & \textbf{17.35} \\
         DeepSeek-V3.1 & 18.37 & 24.49 & 98.98 & 6.12 \\
    \midrule
    \addlinespace[1em] 
        \multicolumn{5}{l}{\textbf{\circled{2} College $\rightarrow$ Sr. Res., Generated (n=459)}} \\
    \midrule
         GPT-5.1 & 1.96 & 26.36 & 99.56 & 0.87 \\
         GPT-5 mini & 3.92 & 5.88 & 69.93 & 0.22 \\
        Sonnet 4.5 & 15.03 & \textbf{32.90} & \textbf{100.0} & \textbf{5.45} \\
        \makecell{ + Thinking} & \textbf{31.59} & 28.76 & 83.66 & 4.79 \\
         DeepSeek-V3.1 & 7.19 & 7.41 & 98.47 & 0.44 \\
    \midrule
    \addlinespace[1em]
        \multicolumn{5}{l}{\textbf{\circled{3} Child $\rightarrow$ Expert, Expert-written (n=98)}} \\
    \midrule
        GPT-5.1 & 18.37 & \textbf{51.02} & \textbf{100.0} & 11.22 \\
         GPT-5 mini & 26.53 & 35.71 & 95.92 & 11.22 \\
        Sonnet 4.5 & 60.20 & 45.92 & \textbf{100.0} & \textbf{33.67} \\
        \makecell{ + Thinking} & \textbf{79.59} & 40.82 & 93.88 & 31.63 \\
         DeepSeek-V3.1 & 46.94 & 36.73 & 97.96 & 16.33 \\
        \bottomrule
    \end{tabular}
    \caption{\textbf{Model performance shown as percent of inputs where measures increase across all $5$ levels.} These percentages represent how often models generate sets of levels that adhere to the desired increase in the complexity measures, so higher is better. The ``All'' column shows the percent of inputs for which the model increases all three measures. 
    }
    \label{tab:model_performance_good_combined}
\end{table}

\subsection{Models inconsistently increase complexity measures}
\label{sec:result_inconsistent}
\begin{figure}[hbt]
    \centering
    \includegraphics[width=\linewidth]{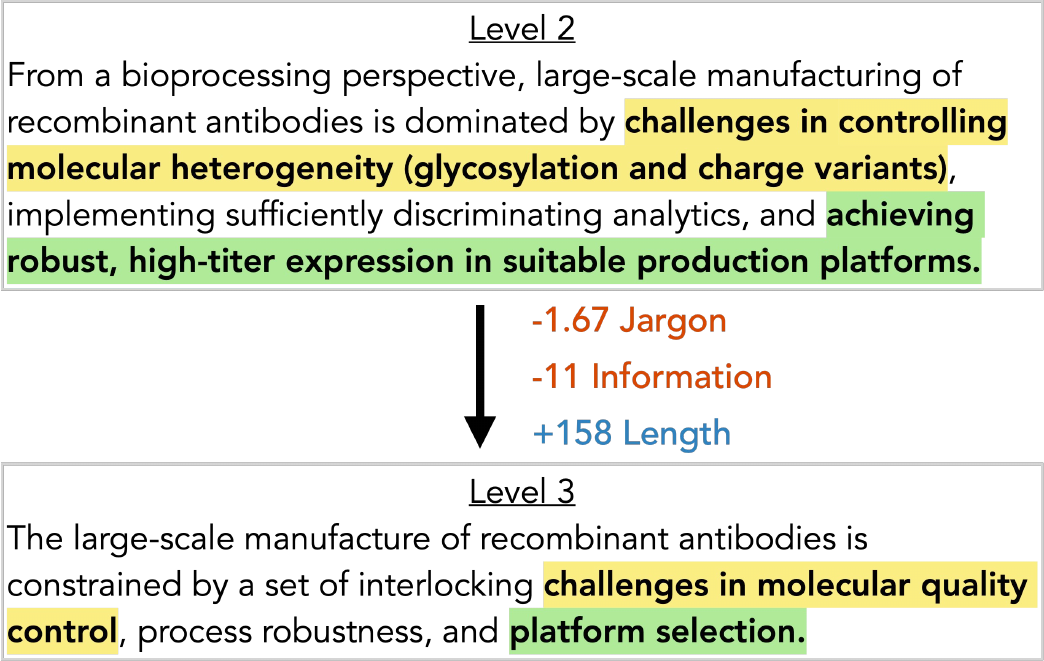}
    \caption{\textbf{Example of text incorrectly decreasing in complexity} Shown are two snippets of text generated by GPT-5.1 that are meant to increase in complexity from Level 2 to Level 3. However, we qualitatively observe that the complexity decreases between analogous phrases (e.g., ``platform selection'' from Level 3 reads simpler than ``achieving robust, high-titer expression in suitable production platforms'' from Level 2). Decreases in \textsc{Jargon} and \textsc{Information} reflect this observation; the measures represent the texts that the snippets are from, not the snippets in isolation, explaining why \textsc{Length} increases.\looseness=-1
    }
    \label{fig:text_example}
\end{figure}

Models often increase the complexity for some inputs while decreasing the complexity for others for the same transition.
In Fig. \ref{fig:boxplots}, this can be seen where distributions cover positive and negative changes for the same model and transition, particularly in \textsc{Jargon} and \textsc{Information}.
For example, when transitioning from Level 3 to 4, Claude Sonnet 4.5 + Thinking increases all three measures for only $58.16\%$ of the expert-written inputs ($48.58\%$ for model-generated inputs) (Tab. \ref{tab:model_performance_percentages}).
Fig. \ref{fig:text_example} shows an example where \textsc{Jargon} and \textsc{Information} both decrease when the complexity is supposed to increase.
\textsc{Length} often moves in the correct direction; however, as shown by our criterion, all measures should move in the correct direction together (discussed further in Sec. \ref{sec:result_length}).\looseness=-1

When considering all transitions (Tab. \ref{tab:model_performance_good_combined}), Claude Sonnet 4.5 + Thinking is the best-performing but only consistently increases complexity measures $17.35\%$ of the time ($4.79\%$ for generated inputs).
Our significance tests further show that for \textsc{Jargon} and \textsc{Information}, the mean of changes moving in the wrong direction is often significantly different from those moving in the right direction ($p<0.01$ for expert-written and generated inputs), implying responses that fail to increase in complexity instead significantly decrease in complexity (as opposed to having no significant difference) (Appendix \ref{sec:sigtest_app}).

The proportion of stagnant changes are at most $1.02\%$ for \textsc{Jargon}, $7.14\%$ for \textsc{Information}, and $0\%$ for \textsc{Length}; thus we focus the proportion of changes that decrease as a stronger indication of complexity going in the wrong direction.

\subsection{Models increase length with complexity}
\label{sec:result_length}
\begin{figure}[hbt]
    \centering
    \includegraphics[width=\linewidth]{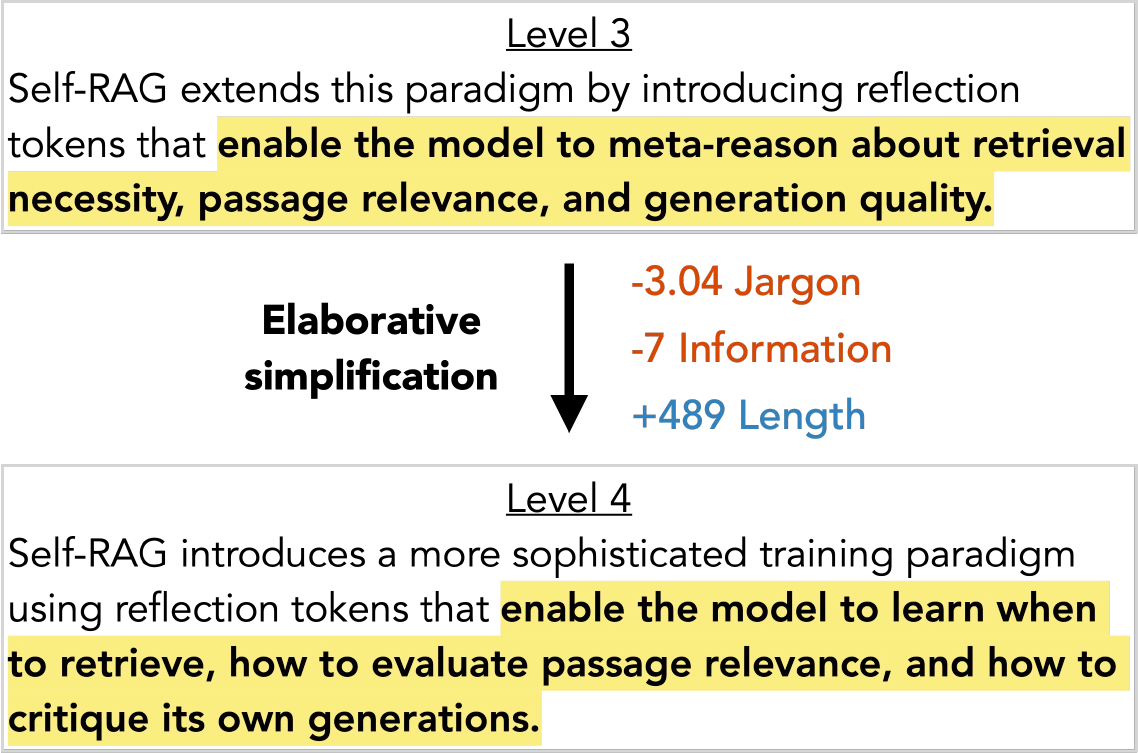}
    \caption{\textbf{Example of elaborative simplification} Snippets of responses generated by Claude Sonnet 4.5 that are supposed to increase in complexity. Between the two, \textsc{Length} increases, while \textsc{Jargon} and \textsc{Information} decrease. The additional text in the Level 4 snippet explains in simple language the term ``meta-reason'' in the Level 3 snippet.}
    \label{fig:elaborative_example}
\end{figure}

Unlike \textsc{Jargon} and \textsc{Information}, models generally increase \textsc{Length} with increasing complexity.
This can be seen in Tab. \ref{tab:model_performance_good_combined} as all models increase \textsc{Length} across the $5$ levels for the majority of their inputs (e.g., DeepSeek-V3.1 increases \textsc{Length} across all levels $98.98\%$ of the time, $98.47\%$ for generated inputs), reflecting the per-transition performance in Tab. \ref{tab:model_performance_percentages}.
However, an increase in \textsc{Length} alone may not always indicate an increase in complexity.
When directly examining responses, we notice that cases where \textsc{Length} increases on its own (without increasing \textsc{Jargon} or \textsc{Information}) can be indicative of elaborative simplification \cite{wu2023elaborative}; an example is shown in Fig. \ref{fig:elaborative_example}.\looseness=-1

\subsection{Models struggle to differentiate responses in later audience levels}
\label{sec:result_later}
In addition to the overall performance discussed in the previous two findings, we consider how the performance varies by transition.
We find that models increase complexity measures more often when transitioning from Level 1  to Level 2 than for the later transitions.
This can be seen in Tab.~\ref{tab:model_performance_percentages} where most models have the highest percentage of inputs moving in the correct direction of complexity in the ``1 to 2'' row compared to the later rows.
For example, GPT-5 mini correctly increases all measures for $62.25\%$ of the inputs when going from Level 1 to 2, which is higher than $42.86\%$, $27.55\%$, and $41.84\%$ for the later transitions.
\subsection{Expanding audience labels can improve performance but exhibits the same trends}
\label{sec:results_ext}
We chose the audiences ``College student'' through ``Senior researcher'' because the wording of the queries in ScholarQABench implies that the inquirer has at least a college education.
We investigate if this choice of audience labels affects the similarity between levels by running an identical evaluation to the first using a different set of audience labels.
Specifically, we use labels from a popular video series called ``5 Levels'' by WIRED\footnote{https://www.wired.com/video/series/5-levels/} that focuses on communicating specialized concepts to $5$ different audiences: Child, Teen, College student, Grad student, and Expert.

While complexity measures increase more often (i.e., the percentages of responses moving in the correct direction are generally higher than those in the first evaluation, Tab.~\ref{tab:model_performance_good_combined}), the same trend of measures moving in the wrong direction holds.
For example, the percent of inputs that Claude Sonnet 4.5 increases in complexity across all transitions improved from $14.29\%$ to $33.67\%$ with the change in audience labels (Tab. \ref{tab:model_performance_good_combined}); however, $33.67\%$ remains relatively low.\looseness=-1

In terms of per-transition performance (Tab. \ref{tab:model_performance_percentages}), the ``Child $\rightarrow$ Expert'' labels increase complexity measures more often than the ``College $\rightarrow$ Senior researcher'' labels in the first evaluation, suggesting that the model interprets the former as more distinct from each other.
However, we observe the same trend where models differentiate between Levels 1 and 2 better than the later levels.
This can be seen when the performance of GPT-5 mini decreases from $80.61\%$ in the first transition to $56.12\%$ in the last (Tab. \ref{tab:model_performance_percentages}).
Similar to the first evaluation, our significance tests further imply that for \textsc{Jargon} and \textsc{Information}, responses that fail to increase in complexity often significantly decrease in complexity ($p<0.01$ for expert-written and generated inputs) (Appendix \ref{sec:sigtest_app}).\looseness=-1

\section{Discussion \& Conclusion}

Enabling users to directly adjust the language of model responses allows LLM-powered systems to better accommodate user needs. While prompting remains the default for adjusting model responses, new malleable interfaces promise more direct user control beyond articulating information needs through language \citep{min2025malleable, zhang2026wordswidgetscontrollablellm}. 
In this paper, we propose an evaluation approach for these settings that tests multiple responses relative to each other rather than single responses in isolation. 
Through a formative user study and additional human validation, we establish that these interactive use cases can be desirable, specifically for controlling language complexity, and identify measures and a criterion for distinguishing between levels of complexity.
To investigate how current models fulfill or stray from users' expectations, we evaluate model-generated levels of complexity for answers to a dataset of scientific queries, varying the answer source and range of intended audiences. We show that evaluating models across levels of complexity reveals model weaknesses that would be difficult to identify in single-response evaluations.
Specifically, models do not consistently increase complexity and that their performance can worsen for higher complexity levels.
Further, while models tend to increase length with complexity levels, they frequently decrease jargon and information, suggesting that models often neglect attributes that are important for perceived complexity and end-user control.\looseness=-1

Our evaluation approach has parallels with other recent trends in model evaluations, such as evaluating multi-turn conversations \citep{laban2026llms}, integrated or live use settings \citep{mehta2026enterprisebenchcorecrafttraininggeneralizable, bragg2026astabenchrigorousbenchmarkingai}, and performance measures based on different intended audiences \citep{joshi2025eliwhy}. 
More generally, as models are integrated into new interfaces \citep[e.g., reading or writing interfaces,][]{10.1145/3772318.3790786, 10.1145/3613904.3642697}, evaluations must shift to incorporate interface-specific criteria.\looseness=-1

\section*{Limitations}
\label{sec:limitations}


We evaluated models using linguistic measures that were strongly supported by prior work and our formative study.
However, there are other potential measures for complexity that could go beyond linguistic characteristics towards the content of the text, such as elaborative simplification and analogies.
Thus, one avenue for future work would be to quantify and evaluate the impact of these additional measures, potentially with additional open-source LLMs as well to increase accessibility.\looseness=-1


We structure our user study around the target setting of people learning unfamiliar scientific topics. While sample sizes for formative studies are often small \cite{guo-etal-2024-personalized,kim2023hmhta,Sauro_Lewis_2012}, we noticed the same salient comments and interactions among our participants, suggesting data saturation for readers in unfamiliar STEM fields. We recognize this is a focused subset of the broader audience of scientific papers; an interesting expansion would be to analyze text from other disciplines.\looseness=-1

Lastly, there are also other options for interactivity to explore.
For our evaluation, the model generates a full response for each of the five levels, and we test different prompting techniques for this (Appendix \ref{sec:prompt}).
However, one could imagine, for example, an interface where users select a specific snippet of text to be rewritten into a different complexity, which leads to other possibilities for direct manipulation.\looseness=-1

\section*{Ethical Considerations}
This work uses LLMs to generate and evaluate responses.
In addition to their impact on the environment \citep{ren2024env,Desislavov2023env}, LLMs can hallucinate and exhibit biases that affect the information they generate \citep{narayanan2024audit,Sharma2024Echo,zhou2025veracity}.
At the same time, we believe that the contribution of this work towards making information from knowledge-intensive domains accessible to people of varying expertise is valuable.
Additionally, we focus on English text; however, this does not necessarily account for contexts with ESL learners who may have different experiences that impact scientific information seeking.


\bibliography{references}

\appendix

\clearpage
\section{Appendix}
\label{sec:appendix}

\subsection{Prompt for Interactive Complexity}
\label{sec:prompt}
\sethlcolor{orange!20}
We use $5$ levels because it allows for some nuance between the versions of text, and a popular video series called ``5 Levels'' by WIRED\footnote{https://www.wired.com/video/series/5-levels/} focuses on communicating specialized concepts to $5$ different audiences which aligns well with the motivation for this work.
To populate the slider with $5$ responses for the user study and run the model evaluation, we experimented with different characteristics of prompts, with \hl{highlights} indicating what worked better: \hl{specifying audience} vs. generic levels, \hl{single} vs. multi-prompt, and defining endpoints of complexity vs. \hl{defining all 5 complexity levels}.
We include example results for our choice between single vs. multi-prompt in Tab. \ref{tab:multiprompt}.
Our final prompt for the model evaluation is shown in Fig. \ref{fig:prompt}; we allowed the model to include a ``References'' section in its responses during the user study.
We also enforced a JSON schema.\looseness=-1

\renewcommand{\arraystretch}{1}
\begin{table}[h]
    \centering
    \small
    \begin{tabular}{l|c|c|c|c}
    \multicolumn{4}{l}{\textbf{Jargon}} & \multicolumn{1}{l}{} \\
    \midrule
    \textbf{Model} & \textbf{1 to 2} & \textbf{2 to 3} & \textbf{3 to 4} & \textbf{4 to 5} \\
    \midrule
         GPT-5.1 & 70.10 & 50.52 & 61.86 & 63.92 \\
         GPT-5 mini & 49.48 & 80.41 & 44.33 & 67.01 \\
         Claude Sonnet 4.5 & 96.91 & 76.29 & 47.42 & 51.55 \\
        \makecell{ + Thinking} & 96.91 & 85.57 & 49.48 & 44.33 \\
         DeepSeek-V3.1 & 67.01 & 72.16 & 57.73 & 53.61 \\
    \midrule
    \addlinespace[1em]
        \multicolumn{4}{l}{\textbf{Information}} & \multicolumn{1}{l}{} \\
    \midrule
         GPT-5.1 & 69.07 & 59.79 & 57.73 & 43.30 \\
         GPT-5 mini & 85.57 & 64.95 & 44.33 & 43.30 \\
         Claude Sonnet 4.5 & 65.98 & 43.30 & 45.36 & 32.99 \\
        \makecell{ + Thinking} & 67.01 & 28.87 & 64.95 & 25.77 \\
         DeepSeek-V3.1 & 54.64 & 50.52 & 44.33 & 41.24 \\
    \midrule
    \addlinespace[1em]
        \multicolumn{4}{l}{\textbf{Length}} & \multicolumn{1}{l}{} \\
    \midrule
        GPT-5.1 & 89.69 & 67.01 & 54.64 & 50.52 \\
         GPT-5 mini & 97.94 & 82.47 & 31.96 & 44.33 \\
         Claude Sonnet 4.5 & 73.20 & 57.73 & 54.64 & 14.43 \\
        \makecell{ + Thinking} & 85.57 & 24.74 & 57.73 & 18.56 \\
         DeepSeek-V3.1 & 63.92 & 59.79 & 49.48 & 44.33 \\
        \bottomrule
    \end{tabular}
    \caption{\textbf{Model performance for College $\rightarrow$ Sr. Res. (Expert-written Inputs, n=98) for multi-prompt approach} To test a multi-prompt approach, we ran an evaluation (n=98) that prompts for each of the levels (College student through Senior researcher) in separate calls (all else the same) and found that it performs worse than the single prompt approach (Tab. \ref{tab:model_performance_percentages}). Results are shown as the percent of inputs moving in the correct direction at each transition.}
    \label{tab:multiprompt}
\end{table}

We chose the audience labels to range from College student to Senior Researcher because the wording of the questions in ScholarQABench implied that the inquirer had at least a college education.
However, as mentioned in Sec.~\ref{sec:results_ext}, we tested using the audiences from the WIRED 5 levels series (Child, Teen, College student, Grad student, Expert) to check the effect of the audience labels, using the same prompt as a template.

\begin{figure*}[!htbp]
    \centering
    
    \begin{tcolorbox}[
    taggingPrompt,
    title={\textbf{College student to Senior researcher Prompt}},
    ]
    \small
        You are given a user query and a report responding to that query as input. \\

        Using information only from the report and query, rewrite the chatbot response into 5 versions where each version responds with a level of complexity appropriate for a College student, Junior Ph.D. student, Senior Ph.D. student, Postdoctoral researcher, or Senior researcher respectively. That is, Version 1 should be written with a level of complexity that a college student would understand, Version 2 should be written with a level of complexity that a junior Ph.D. student would understand, and so on. Assume that the user understands the words in their query. \\
    
        Preserve mentions of papers and citations by preserving abbreviated, in-text citations. To be clear, do not write out a References section, just use in-text citations like this: (Vaswani, 2017). \\
    
        Do not add any additional text like greetings or ornamental words.
        
        
    \end{tcolorbox}
    
    \caption{\textbf{Prompt for Interactive Complexity}}
    \label{fig:prompt}
\end{figure*}

\subsection{Additional Data Analysis and Information}
\subsubsection{Queries from ScholarQABench}
\label{sec:scholarqabench_app}
For the first evaluation, we use the queries from ScholarQA-Multi, a subset of ScholarQABench~\citep{asai2024openscholar}\footnote{https://github.com/AkariAsai/ScholarQABench} that contains $98$ expert-written scientific queries and answers across multiple STEM fields.
The distribution of queries across fields, along with sample queries, is shown in Table \ref{tab:data_distribution}.

\renewcommand{\arraystretch}{1.5}
\begin{table*}[bt]
    \centering
    \small
    \begin{tabular}{l|l|l}
    \toprule
         \textbf{Discipline} & \textbf{\% of Data} & \textbf{Sample Query} \\
         \midrule
         Biology & 20.4 & \textit{What are the biochemical analytical tools to assess the integrity and stability of LNPs?} \\ 
         Biophysics & 9.2 & \textit{Find some papers that discuss the methods to suppress multiple light scattering effect.} \\ 
         Computer Science & 33.7 & \textit{Could you please provide some references to work on multi-document summarization?} \\ 
         Photonics & 30.6 & \textit{What progress has been made in trapping and controlling multiple nanoparticles?} \\ 
         Physics & 6.1 & \textit{What are the ways to perform optomechanical cooling?} \\
         \bottomrule
    \end{tabular}
    \caption{\textbf{Distribution of data across disciplines} ScholarQA-Multi consists of $98$ queries distributed across Biology, Biophysics, Computer Science, Photonics, and Physics; a sample question from each domain is shown.}
    \label{tab:data_distribution}
\end{table*}

\subsubsection{Descriptive Statistics of Boxplots}
\label{sec:descriptive_stats_app}
This section contains descriptive statistics, specifically the means and standard deviations, for the distributions shown in Fig. \ref{fig:boxplots}.
Tables \ref{tab:descriptive_stats}, \ref{tab:descriptive_stats_expanded}, and \ref{tab:descriptive_stats_wired} are for the ``College $\rightarrow$ Sr. Res., Expert-written Inputs (n=98)'', ``College $\rightarrow$ Sr. Res., Generated Inputs (n=459)'', and ``Child $\rightarrow$ Expert, Expert-written Inputs (n=98)'' groups respectively.

\renewcommand{\arraystretch}{1}
\begin{table*}[t]
    \centering
    \small
    \begin{tabular}{c|l|c|c|c|}
        & \textbf{Model} & \textbf{Jargon} & \textbf{Info.} & \textbf{Length} \\
    \midrule
         & GPT-5.1 & 6.36 ± 3.89 & 8.31 ± 8.56 & 1314.48 ± 715.12 \\
         \multirow{1}{*}[-5pt]{\rotatebox{90}{\textbf{1 to 2}}} &GPT-5 mini & 6.01 ± 4.79 & 3.39 ± 5.55 & 355.53 ± 181.59 \\
         &Claude Sonnet 4.5 & 15.31 ± 5.33 & 5.13 ± 5.63 & 532.77 ± 206.22 \\
         &\makecell{ + Thinking} & 14.21 ± 5.65 & 4.73 ± 5.64 & 455.89 ± 185.6 \\
         &DeepSeek-V3.1 & 13.2 ± 5.5 & 3.06 ± 3.62 & 311.71 ± 149.61 \\

    \midrule
         & GPT-5.1 & 0.21 ± 2.55 & 9.02 ± 8.97 & 1280.07 ± 552.53 \\
         \multirow{1}{*}[-5pt]{\rotatebox{90}{\textbf{2 to 3}}} &GPT-5 mini & 0.63 ± 3.45 & 3.98 ± 6.29 & 404.54 ± 269.97 \\
         &Claude Sonnet 4.5 & 2.79 ± 3.27 & 5.88 ± 5.9 & 515.05 ± 267.77 \\
         &\makecell{ + Thinking} & 3.53 ± 3.0 & 4.32 ± 5.92 & 459.96 ± 237.09 \\
         &DeepSeek-V3.1 & 1.91 ± 3.4 & 3.57 ± 4.85 & 272.45 ± 158.17 \\

    \midrule
         & GPT-5.1 & 0.00 ± 1.91 & 4.78 ± 10.46 & 1015.32 ± 522.71 \\
         \multirow{1}{*}[-5pt]{\rotatebox{90}{\textbf{3 to 4}}} &GPT-5 mini & 1.16 ± 3.07 & 2.04 ± 6.57 & 177.31 ± 243.21 \\
         &Claude Sonnet 4.5 & 1.69 ± 2.77 & 5.88 ± 7.41 & 681.84 ± 340.44 \\
         &\makecell{ + Thinking} & 1.72 ± 2.7 & 5.69 ± 7.59 & 547.86 ± 299.72 \\
         &DeepSeek-V3.1 & 0.44 ± 3.82 & 1.48 ± 4.83 & 221.73 ± 121.54 \\

    \midrule
         & GPT-5.1 & -0.73 ± 1.62 & 6.77 ± 9.88 & 1228.54 ± 602.67 \\
         \multirow{1}{*}[-5pt]{\rotatebox{90}{\textbf{4 to 5}}} &GPT-5 mini & 0.94 ± 3.03 & 3.79 ± 6.56 & 509.57 ± 405.4 \\
         &Claude Sonnet 4.5 & 0.55 ± 2.27 & 8.49 ± 8.96 & 1098.28 ± 703.87 \\
         &\makecell{ + Thinking} & 1.61 ± 2.49 & 5.14 ± 7.66 & 661.05 ± 722.9 \\
         &DeepSeek-V3.1 & 0.68 ± 2.86 & 3.3 ± 4.5 & 351.92 ± 164.12 \\

    \bottomrule
    \end{tabular}
    \caption{\textbf{Model performance per transition for College $\rightarrow$ Sr. Res. (Expert-written Inputs, n=98)} Mean and standard deviation are shown.}
    \label{tab:descriptive_stats}
\end{table*}

\begin{table*}[t]
    \centering
    \small
    \begin{tabular}{c|l|c|c|c|}
        & \textbf{Model} & \textbf{Jargon} & \textbf{Info.} & \textbf{Length} \\
    \midrule
         & GPT-5.1 & 6.02 ± 4.21 & 11.27 ± 12.86 & 1629.2 ± 853.56 \\
         \multirow{1}{*}[-5pt]{\rotatebox{90}{\textbf{1 to 2}}} &GPT-5 mini & 5.1 ± 4.92 & 1.62 ± 6.66 & 208.1 ± 171.21 \\
         &Claude Sonnet 4.5 & 17.74 ± 5.67 & 4.58 ± 6.59 & 457.04 ± 224.88 \\
         &\makecell{ + Thinking} & 14.56 ± 5.18 & 5.97 ± 7.53 & 491.03 ± 291.95 \\
         &DeepSeek-V3.1 & 16.99 ± 6.88 & 2.72 ± 4.5 & 238.97 ± 115.29 \\

    \midrule
         & GPT-5.1 & -0.27 ± 2.45 & 10.15 ± 14.89 & 1664.0 ± 781.91 \\
         \multirow{1}{*}[-5pt]{\rotatebox{90}{\textbf{2 to 3}}} &GPT-5 mini & 0.24 ± 3.58 & 3.45 ± 8.34 & 326.81 ± 186.53 \\
         &Claude Sonnet 4.5 & 2.1 ± 3.13 & 5.8 ± 7.99 & 625.99 ± 446.66 \\
         &\makecell{ + Thinking} & 2.52 ± 3.13 & 5.42 ± 7.71 & 691.07 ± 488.99 \\
         &DeepSeek-V3.1 & 1.45 ± 4.59 & 2.03 ± 4.85 & 211.61 ± 126.37 \\

    \midrule
         & GPT-5.1 & -0.64 ± 1.8 & 5.27 ± 15.2 & 1195.75 ± 605.11 \\
         \multirow{1}{*}[-5pt]{\rotatebox{90}{\textbf{3 to 4}}} &GPT-5 mini & 0.09 ± 3.64 & -0.14 ± 8.83 & 120.39 ± 189.7 \\
         &Claude Sonnet 4.5 & 0.34 ± 2.54 & 8.54 ± 10.3 & 900.51 ± 648.02 \\
         &\makecell{ + Thinking} & 0.98 ± 2.6 & 6.77 ± 10.06 & 847.88 ± 664.22 \\
         &DeepSeek-V3.1 & -0.28 ± 4.2 & 1.45 ± 5.26 & 214.11 ± 123.19 \\

    \midrule
         & GPT-5.1 & -1.08 ± 1.56 & 5.75 ± 14.88 & 1407.18 ± 797.09 \\
         \multirow{1}{*}[-5pt]{\rotatebox{90}{\textbf{4 to 5}}} &GPT-5 mini & -0.16 ± 3.45 & 3.79 ± 8.55 & 567.35 ± 349.88 \\
         &Claude Sonnet 4.5 & -0.09 ± 2.31 & 9.92 ± 13.48 & 1399.97 ± 976.72 \\
         &\makecell{ + Thinking} & 0.98 ± 2.44 & 4.7 ± 12.34 & 990.22 ± 1165.34 \\
         &DeepSeek-V3.1 & -0.22 ± 3.48 & 2.52 ± 6.15 & 285.56 ± 170.85 \\

    \bottomrule
    \end{tabular}
    \caption{\textbf{Model performance per transition for College $\rightarrow$ Sr. Res. (Generated Inputs, n=459)} Mean and standard deviation are shown.}
    \label{tab:descriptive_stats_expanded}
\end{table*}

\begin{table*}[t]
    \centering
    \small
    \begin{tabular}{c|l|c|c|c|}
        & \textbf{Model} & \textbf{Jargon} & \textbf{Info.} & \textbf{Length} \\
    \midrule
         & GPT-5.1 & 15.94 ± 5.75 & 7.09 ± 6.67 & 719.06 ± 367.96 \\
         \multirow{1}{*}[-5pt]{\rotatebox{90}{\textbf{1 to 2}}} &GPT-5 mini & 17.44 ± 7.19 & 4.81 ± 4.84 & 337.99 ± 161.3 \\
         &Claude Sonnet 4.5 & 19.8 ± 6.31 & 5.57 ± 4.56 & 451.6 ± 134.68 \\
         &\makecell{ + Thinking} & 19.01 ± 6.45 & 6.4 ± 5.24 & 463.13 ± 154.94 \\
         &DeepSeek-V3.1 & 14.26 ± 6.82 & 3.64 ± 3.38 & 223.52 ± 99.24 \\

    \midrule
         & GPT-5.1 & 6.43 ± 3.99 & 8.48 ± 6.39 & 1081.16 ± 509.61 \\
         \multirow{1}{*}[-5pt]{\rotatebox{90}{\textbf{2 to 3}}} &GPT-5 mini & 2.85 ± 3.81 & 4.05 ± 4.96 & 410.45 ± 194.22 \\
         &Claude Sonnet 4.5 & 11.09 ± 5.52 & 5.15 ± 5.71 & 511.41 ± 210.74 \\
         &\makecell{ + Thinking} & 10.73 ± 5.09 & 4.37 ± 5.65 & 529.91 ± 236.12 \\
         &DeepSeek-V3.1 & 16.23 ± 6.13 & 3.36 ± 3.49 & 340.62 ± 139.68 \\

    \midrule
         & GPT-5.1 & 0.43 ± 2.72 & 8.83 ± 9.52 & 1371.26 ± 572.88 \\
         \multirow{1}{*}[-5pt]{\rotatebox{90}{\textbf{3 to 4}}} &GPT-5 mini & 1.5 ± 3.43 & 4.1 ± 5.76 & 400.8 ± 185.05 \\
         &Claude Sonnet 4.5 & 3.47 ± 3.4 & 5.65 ± 5.57 & 625.2 ± 316.5 \\
         &\makecell{ + Thinking} & 4.28 ± 3.15 & 6.84 ± 6.7 & 678.54 ± 372.7 \\
         &DeepSeek-V3.1 & 3.06 ± 3.52 & 2.94 ± 4.36 & 333.94 ± 146.99 \\

    \midrule
         & GPT-5.1 & -0.79 ± 1.91 & 12.46 ± 11.68 & 2226.66 ± 816.5 \\
         \multirow{1}{*}[-5pt]{\rotatebox{90}{\textbf{4 to 5}}} &GPT-5 mini & 1.63 ± 2.91 & 5.19 ± 6.64 & 669.51 ± 359.69 \\
         &Claude Sonnet 4.5 & 2.2 ± 2.8 & 7.86 ± 7.53 & 829.68 ± 515.69 \\
         &\makecell{ + Thinking} & 3.44 ± 2.7 & 5.51 ± 7.87 & 686.22 ± 527.81 \\
         &DeepSeek-V3.1 & 1.13 ± 3.4 & 4.28 ± 4.58 & 378.29 ± 144.78 \\

    \bottomrule
    \end{tabular}
    \caption{\textbf{Model performance per transition for Child $\rightarrow$ Expert (Expert-written Inputs, n=98)} Mean and standard deviation are shown.}
    \label{tab:descriptive_stats_wired}
\end{table*}

\subsubsection{Significance Testing of Results}
\label{sec:sigtest_app}
As mentioned in Sec. \ref{sec:model_eval_results}, to test if the distributions of changes going in the wrong direction are significantly different from the distributions of changes going in the right direction (Fig. \ref{fig:boxplots}), we run Welch's $t$-tests, applying the Bonferroni correction, comparing those distributions at each of the 4 transitions.
That is, at each of the 4 transitions (Level 1 to 2, Level 2 to 3, etc.), we compare whether or not the magnitudes of the changes in the wrong direction are significantly different from the magnitudes of the changes in the right direction.
We report the $95\%$ confidence intervals with their significance for each of the evaluations in Tables \ref{tab:sigtest}, \ref{tab:sigtest_expanded}, and \ref{tab:sigtest_wired} respectively.
For all models, we find that when there are more than a few inputs for which the model moves the complexity measure towards the incorrect direction, the change often moves significantly in the wrong direction (i.e., when the model incorrectly decreases \textsc{Jargon}, it does so to a significant degree).\looseness=-1
\renewcommand{\arraystretch}{1}
\begin{table*}[h]
    \centering
    \small
    \begin{tabular}{l|c|c|c|c}
        \multicolumn{4}{l}{\textbf{Jargon}} & \multicolumn{1}{l}{} \\
        \midrule
            Model & 1 to 2 & 2 to 3 & 3 to 4 & 4 to 5 \\
        \midrule
            GPT-5.1 & [7.67,12.36]** & [3.58,4.77]** & [2.57,3.53]** & [1.77,2.84]** \\
            GPT-5 mini & [8.2,11.16]** & [4.75,6.4]** & [3.99,5.57]** & [3.69,5.35]** \\
            Claude Sonnet 4.5 & n/a & [4.16,6.03]** & [4.25,5.55]** & [3.14,4.21]** \\
            \makecell{ + Thinking} & n/a & [4.58,6.54]** & [3.86,5.58]** & [3.66,5.06]** \\
            DeepSeek-V3.1 & n/a & [5.12,6.62]** & [5.05,6.96]** & [4.01,5.41]** \\
        \midrule
        \addlinespace[1em]
            \multicolumn{4}{l}{\textbf{Information}} & \multicolumn{1}{l}{} \\
        \midrule
            GPT-5.1 & [12.21,16.03]** & [13.19,17.18]** & [14.05,20.62]** & [13.2,18.78]** \\
            GPT-5 mini & [7.55,10.39]** & [9.07,12.79]** & [8.66,11.9]** & [9.63,12.91]** \\
            Claude Sonnet 4.5 & [7.77,11.23]** & [8.41,11.27]** & [10.01,13.92]** & [12.72,19.38]** \\
            \makecell{ + Thinking} & [7.63,10.91]** & [8.47,12.81]** & [9.67,13.36]** & [9.87,14.47]** \\
            DeepSeek-V3.1 & [4.62,6.44]** & [6.63,9.08]** & [5.17,9.09]** & [5.75,7.95]** \\
        \midrule
        \addlinespace[1em]
            \multicolumn{4}{l}{\textbf{Length}} & \multicolumn{1}{l}{} \\
        \midrule
            GPT-5.1 & n/a & n/a & n/a & n/a \\
            GPT-5 mini & n/a & [325.31,608.84]* & [311.7,451.77]** & [227.28,1647.9] \\
            Claude Sonnet 4.5 & n/a & n/a & n/a & n/a \\
            \makecell{ + Thinking} & n/a & n/a & n/a & [1078.36,1734.76]** \\
            DeepSeek-V3.1 & n/a & n/a & n/a & n/a \\
        \midrule
    \end{tabular}
    \caption{\textbf{Significance of difference between incorrect and correct changes in measures for College $\rightarrow$ Sr. Res. (Expert-written Inputs, n=98) evaluation} $95\%$ confidence intervals of the difference between the means are shown. ** indicates p<0.01, * indicates p<0.05, n/a means that no or few inputs went in the wrong direction.}
    \label{tab:sigtest}
\end{table*}

\renewcommand{\arraystretch}{1}
\begin{table*}[h]
    \centering
    \small
    \begin{tabular}{l|c|c|c|c}
        \multicolumn{4}{l}{\textbf{Jargon}} & \multicolumn{1}{l}{} \\
        \midrule
            Model & 1 to 2 & 2 to 3 & 3 to 4 & 4 to 5 \\
        \midrule
            GPT-5.1 & [7.8,9.68]** & [3.4,4.02]** & [2.54,2.97]** & [2.24,2.6]** \\
            GPT-5 mini & [8.22,9.67]** & [5.26,6.06]** & [5.46,6.25]** & [5.03,5.82]** \\
            Claude Sonnet 4.5 & n/a & [4.89,5.66]** & [3.79,4.36]** & [3.34,3.89]** \\
            \makecell{ + Thinking} & n/a & [4.95,5.72]** & [3.89,4.5]** & [3.55,4.17]** \\
            DeepSeek-V3.1 & [-13.56,53.79] & [6.79,7.84]** & [6.26,7.19]** & [5.15,5.93]** \\
        \midrule
        \addlinespace[1em]
            \multicolumn{4}{l}{\textbf{Information}} & \multicolumn{1}{l}{} \\
        \midrule
            GPT-5.1 & [18.18,22.45]** & [20.71,25.76]** & [20.74,24.97]** & [20.95,25.42]** \\
            GPT-5 mini & [9.35,11.0]** & [11.38,13.53]** & [11.88,14.07]** & [12.55,14.7]** \\
            Claude Sonnet 4.5 & [9.51,11.14]** & [11.2,13.27]** & [14.74,17.48]** & [17.95,21.33]** \\
            \makecell{ + Thinking} & [11.19,13.04]** & [11.3,13.3]** & [13.99,16.55]** & [17.45,20.66]** \\
            DeepSeek-V3.1 & [6.43,7.51]** & [6.9,8.05]** & [7.83,9.03]** & [8.28,9.82]** \\
        \midrule
        \addlinespace[1em]
            \multicolumn{4}{l}{\textbf{Length}} & \multicolumn{1}{l}{} \\
        \midrule
            GPT-5.1 & n/a & n/a & n/a & [1416.27,1572.11]** \\
            GPT-5 mini & [261.89,310.97]** & [295.02,489.73]* & [272.72,325.22]** & [513.46,977.82]** \\
            Claude Sonnet 4.5 & n/a & n/a & n/a & n/a \\
            \makecell{ + Thinking} & n/a & n/a & [944.9,1332.41]** & [1648.57,1946.75]** \\
            DeepSeek-V3.1 & [229.77,273.36]** & n/a & [213.65,248.93]** & [219.26,400.55] \\
        \midrule
    \end{tabular}
    \caption{\textbf{Significance of difference between incorrect and correct changes in measures for College $\rightarrow$ Sr. Res. (Generated Inputs, n=459) evaluation} $95\%$ confidence intervals of the difference between the means are shown. ** indicates p<0.01, * indicates p<0.05, n/a means that no or few inputs went in the wrong direction.}
    \label{tab:sigtest_expanded}
\end{table*}

\renewcommand{\arraystretch}{1}
\begin{table*}[h]
    \centering
    \small
    \begin{tabular}{l|c|c|c|c}
        \multicolumn{4}{l}{\textbf{Jargon}} & \multicolumn{1}{l}{} \\
        \midrule
            Model & 1 to 2 & 2 to 3 & 3 to 4 & 4 to 5 \\
        \midrule
            GPT-5.1 & n/a & [6.55,8.54]** & [3.59,5.17]** & [2.7,3.59]** \\
            GPT-5 mini & n/a & [5.69,7.49]** & [4.58,6.33]** & [3.8,5.3]** \\
            Claude Sonnet 4.5 & n/a & [9.59,15.01]* & [5.08,6.7]** & [4.26,5.73]** \\
            \makecell{ + Thinking} & n/a & [10.21,12.76]** & [5.19,6.55]** & [4.11,6.03]** \\
            DeepSeek-V3.1 & [3.89,30.38] & n/a & [5.15,6.96]** & [4.66,6.33]** \\
        \midrule
        \addlinespace[1em]
            \multicolumn{4}{l}{\textbf{Information}} & \multicolumn{1}{l}{} \\
        \midrule
            GPT-5.1 & [9.59,13.23]** & [8.63,15.48]** & [12.76,17.71]** & [16.53,23.04]** \\
            GPT-5 mini & [6.27,8.77]** & [7.31,10.06]** & [6.98,9.94]** & [9.01,13.65]** \\
            Claude Sonnet 4.5 & [7.18,9.88]** & [8.26,12.1]** & [6.86,9.49]** & [10.06,14.1]** \\
            \makecell{ + Thinking} & [7.19,10.83]** & [8.13,11.82]** & [9.34,13.1]** & [11.58,15.35]** \\
            DeepSeek-V3.1 & [5.03,7.64]** & [4.96,7.48]** & [5.46,7.78]** & [6.34,9.27]** \\
        \midrule
        \addlinespace[1em]
            \multicolumn{4}{l}{\textbf{Length}} & \multicolumn{1}{l}{} \\
        \midrule
            GPT-5.1 & n/a & n/a & n/a & n/a \\
            GPT-5 mini & n/a & n/a & n/a & [647.79,913.09]** \\
            Claude Sonnet 4.5 & n/a & n/a & n/a & n/a \\
            \makecell{ + Thinking} & n/a & n/a & n/a & [746.12,1315.55]** \\
            DeepSeek-V3.1 & [177.56,322.69] & n/a & n/a & n/a \\
        \midrule
    \end{tabular}
    \caption{\textbf{Significance of difference between incorrect and correct changes in measures for Child $\rightarrow$ Expert (Expert-written Inputs, n=98) evaluation} $95\%$ confidence intervals of the difference between the means are shown. ** indicates p<0.01, * indicates p<0.05, n/a means that no or few inputs went in the wrong direction.}
    \label{tab:sigtest_wired}
\end{table*}

\subsection{Model Configuration Details}
\label{sec:model_config}
Tab. \ref{tab:configs} shows the model configurations for the model evaluation.
We prompted all but 1 input only once; the 1 input that we had to regenerate was initially refused by Sonnet 4.5, possibly due to the topic.
We ran the evaluation during November 2025 through January 2026.\looseness=-1

\begin{table*}
    \centering
    \small
    \begin{tabular}{l|l}
    \toprule
         Model & Hyperparameters \\
         \midrule
         \texttt{gpt-5.1-2025-11-13} & temperature = 0 \\
         \texttt{gpt-5-mini-2025-08-07} & none specified \\
         \texttt{Claude Sonnet 4.5} & temperature = 0; max\_tokens = 9000 \\
         \texttt{Claude Sonnet 4.5 + Thinking} & max\_tokens = 11192 (includes 3000 for thinking) \\
         \texttt{deepseek.v3-v1:0} & temperature = 0; maxTokens = 9000 \\
         \bottomrule
    \end{tabular}
    \caption{\textbf{Model configurations}}
    \label{tab:configs}
\end{table*}

\subsection{Additional Measures}
\subsubsection{Flesch-Kincaid Reading Ease Score Data}
\label{sec:flesch_data}
The Flesch-Kincaid Reading Ease Score is a commonly-used scale for rating complexity \citep{flesch1948,joshi2025eliwhy,cochrane,farber2025simplifymytext}.
However, past work has shown that Flesch-Kincaid scores do not provide a reliable measure of complexity \citep{cachola2025flesch,tanprasert2021flesch,imperial2023flesch}, and we found that our findings remain the same as the scores exhibit similar trends to \textsc{Jargon} and \textsc{Information}.

\begin{figure}
    \centering
    \includegraphics[width=0.8\linewidth]{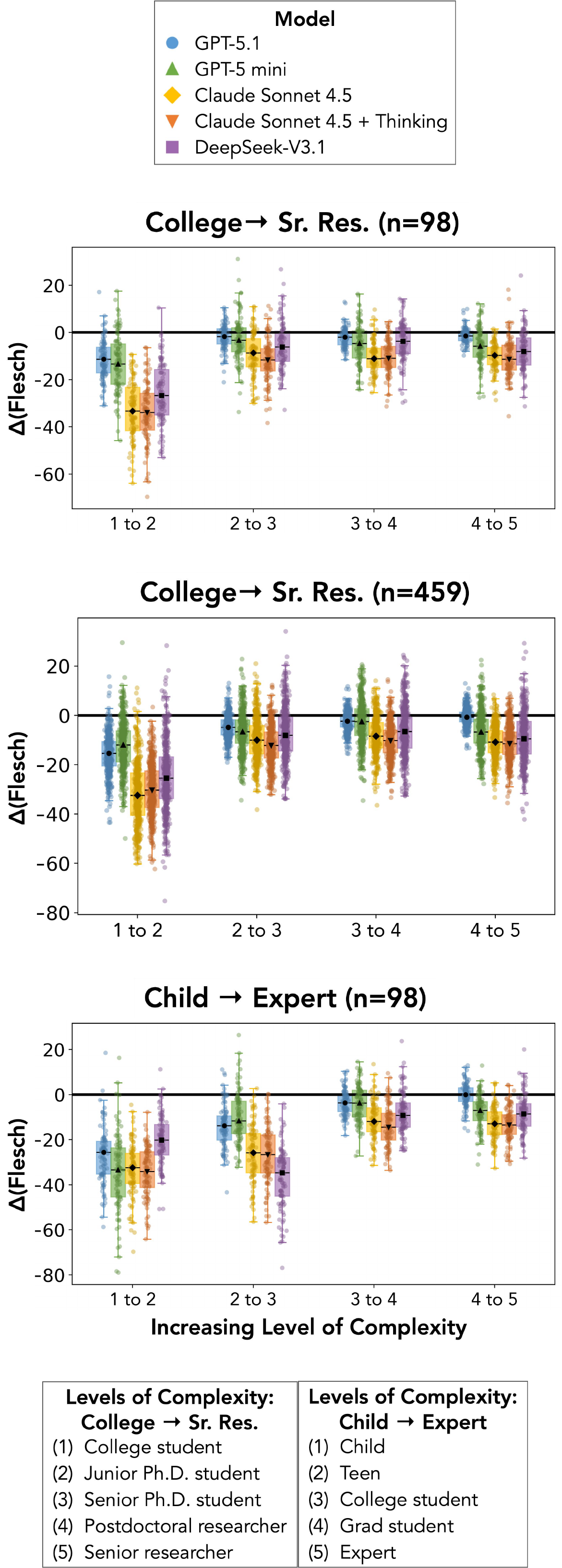}
    \caption{\textbf{Flesch-Kincaid Reading Ease score data} These plots show the distributions of changes in the Flesch-Kincaid scores between consecutive levels for the two sets of audience labels that we prompted with. Expert-written inputs were used for $n=98$, and model-generated inputs were used for $n=459$. Since a higher Flesch-Kincaid score means that the text is more readable, \textit{decreasing} the scores as complexity increases is desirable.}
    \label{fig:flesch_data}
\end{figure}

Fig. \ref{fig:flesch_data} shows the changes in Flesch-Kincaid scores across levels labeled with our initial audiences and the WIRED audiences.
Flesch-Kincaid scores follow the opposite trend to the measures in the main paper; a \textit{decrease} in the score indicates higher complexity.
Similar to \textsc{Jargon} and \textsc{Information} (Fig. \ref{fig:boxplots}), we observe inconsistent increases in complexity (i.e., decreasing Flesch-Kincaid scores).

\subsubsection{Information Density}
\label{sec:density_app}
\textsc{Information} is one of the three measures used in our evaluations.
An alternative to providing the raw number of information points as we do for \textsc{Information} is to calculate the ratio between information and length.
We test such a measure, using the number of sentences to represent length following the idea that a sentence can represent a single information point.
The results are detailed in Tab. \ref{tab:info_density} as the percent of inputs moving in the correct direction at each transition, similar to Tab. \ref{tab:model_performance_percentages}. 
We find that model performance according to this measure is worse than according to the other measures, having fewer inputs moving in the correct direction per transition.
Additionally, while we divide by the number of sentences, there is no standardized way to calculate information density.
\renewcommand{\arraystretch}{1}
\begin{table}[h]
    \centering
    \small
    \begin{tabular}{l|c|c|c|c}
        \textbf{Model} & \textbf{1 to 2} & \textbf{2 to 3} & \textbf{3 to 4} & \textbf{4 to 5} \\
        \midrule
         GPT-5.1 & 26.53 & 39.80 & 38.78 & 40.82  \\
         GPT-5 mini & 47.96 & 42.86 & 47.96 & 44.90 \\
         Claude Sonnet 4.5 & 66.33 & 61.22 & 56.12 & 67.35 \\
         \makecell{+ Thinking} & 68.37 & 63.27 & 66.33 & 63.27 \\
         DeepSeek-V3.1 & 57.14 & 63.27 & 56.12 & 56.12
    \end{tabular}
    \caption{\textbf{Model performance per transition for College $\rightarrow$ Sr. Res. (Expert-written Inputs, n=98) according to information density} Each model's performance is shown as the percent of inputs where information density moves in the correct direction at each transition. A higher percentage means that the model performed better at that transition, by more often increasing complexity.}
    \label{tab:info_density}
\end{table}


\subsection{User Study Details and Materials}
\label{sec:user_study_materials_app}
\subsubsection{Study Materials}
\label{sec:materials}
\begin{figure}
    \centering
    \includegraphics[width=\linewidth]{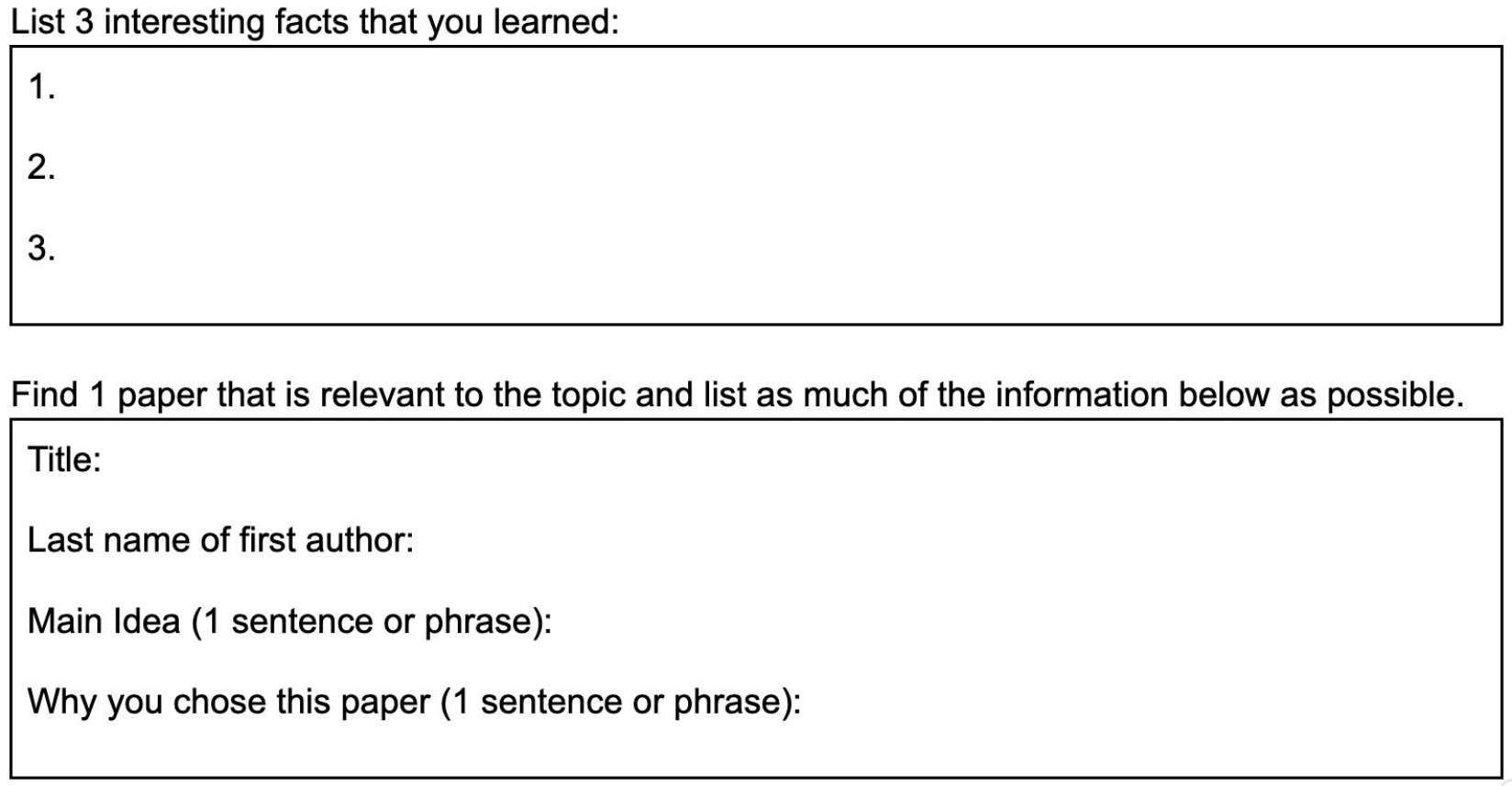}
    \caption{\textbf{Task Questions}}
    \label{fig:task_questions}
\end{figure}

Participants were guided through a 1-hour Zoom session where they interacted with two interfaces, one conventional chat interface and one with interactive complexity as described in Sec. \ref{sec:userstudy_method}.
For the conventional chat interface, participants were told ``\textit{This is a simple chatbot where you ask a question and it provides a response}''.
Since the interactive complexity version had slider features (Fig.~\ref{fig:cui}), participants watched a 1-minute video tutorial.
After learning how to use the interfaces, participants were then given task instructions as shown in Fig.~\ref{fig:task_instructions}.
The task questions are displayed in Fig.~\ref{fig:task_questions}; participants were also provided with a note-taking box in the same document.
After the study, participants received a $\$25$ gift card, which is above the minimum wage policy in the area, and were provided with the disclaimer shown in Fig.~\ref{fig:disclaimer}.
This study was approved by our institution's IRB.

\begin{figure}[!htbp]
    \centering
    
    \begin{tcolorbox}[
    taggingPrompt,
    title={\textbf{Task Instructions Given to Participants}},
    ]
    \small
        Before this session, you provided a topic: [topic provided by participant].
        
        This document contains a few tasks about this topic for you to complete using the chatbot. It also contains an area for you to take notes if you would like to. \\
        
        You will have 15 minutes to complete the tasks using only the chatbot; I will provide you with time warnings. \\
        
        Please focus on using the chatbot rather than reading through links or papers. You can open links to verify content from the system, but do not read the entire paper. \\
        
        While doing the tasks, you will be ``thinking aloud''.
        Basically, we want you to tell us everything that goes through your mind from the start to the end as you complete the tasks.
        When you are thinking aloud:\\
        - There are no right or wrong answers. You are not being tested. \\
        - Be honest. If you feel confused, frustrated, or surprised, please say so.\\
        - Keep talking. Demonstrate your entire process, what you’re doing, why you’re doing it, what you’re thinking, and what you're expecting \\
        - Focus on the task. Complete the tasks to the best of your ability, while simultaneously describing your process. \\
        For example, if you click something, you could say ``I am clicking this because'' followed by your reason. If you’re confused, you could say ``I’m not sure what to do here because'' again followed by a reason.
    \end{tcolorbox}
    
    \caption{\textbf{Task Instructions Given to Participants}}
    \label{fig:task_instructions}
\end{figure}

\begin{figure}[!htbp]
    \centering
    
    \begin{tcolorbox}[
    taggingPrompt,
    title={\textbf{Post-Study Disclaimer}},
    ]
    \small
        To create a realistic setting, we showed AI answers that are directly from responses from an actual AI system. As known, AI systems can make up information. Please note that the AI answers you saw in this study may have been inaccurate, incomplete, or inconsistent, even when they sounded convincing.
    \end{tcolorbox}
    
    \caption{\textbf{Post-Study Disclaimer}}
    \label{fig:disclaimer}
\end{figure}

\subsubsection{User Study Interface}
We extended a chat interface from the Streamlit framework\footnote{https://docs.streamlit.io/develop/tutorials/chat-and-llm-apps/build-conversational-apps} to build our conventional and interactive interfaces.
More details are provided in Sec.~\ref{sec:userstudy_method}.

\subsubsection{Complexity of Responses in User Study}
\label{sec:user_study_complexity}
In this section, we report the changes in complexity for the aggregated $45$ responses across the $16$ participants from the interactive condition of the user study.
We use GPT-5 mini to supply the responses as a low-latency option.
The distributions generally match those of GPT-5 mini in the model evaluation (Fig. \ref{fig:boxplots}), confirming that the user study was a fair instantiation of the model evaluation.

\begin{figure}
    \centering
    \includegraphics[width=\linewidth]{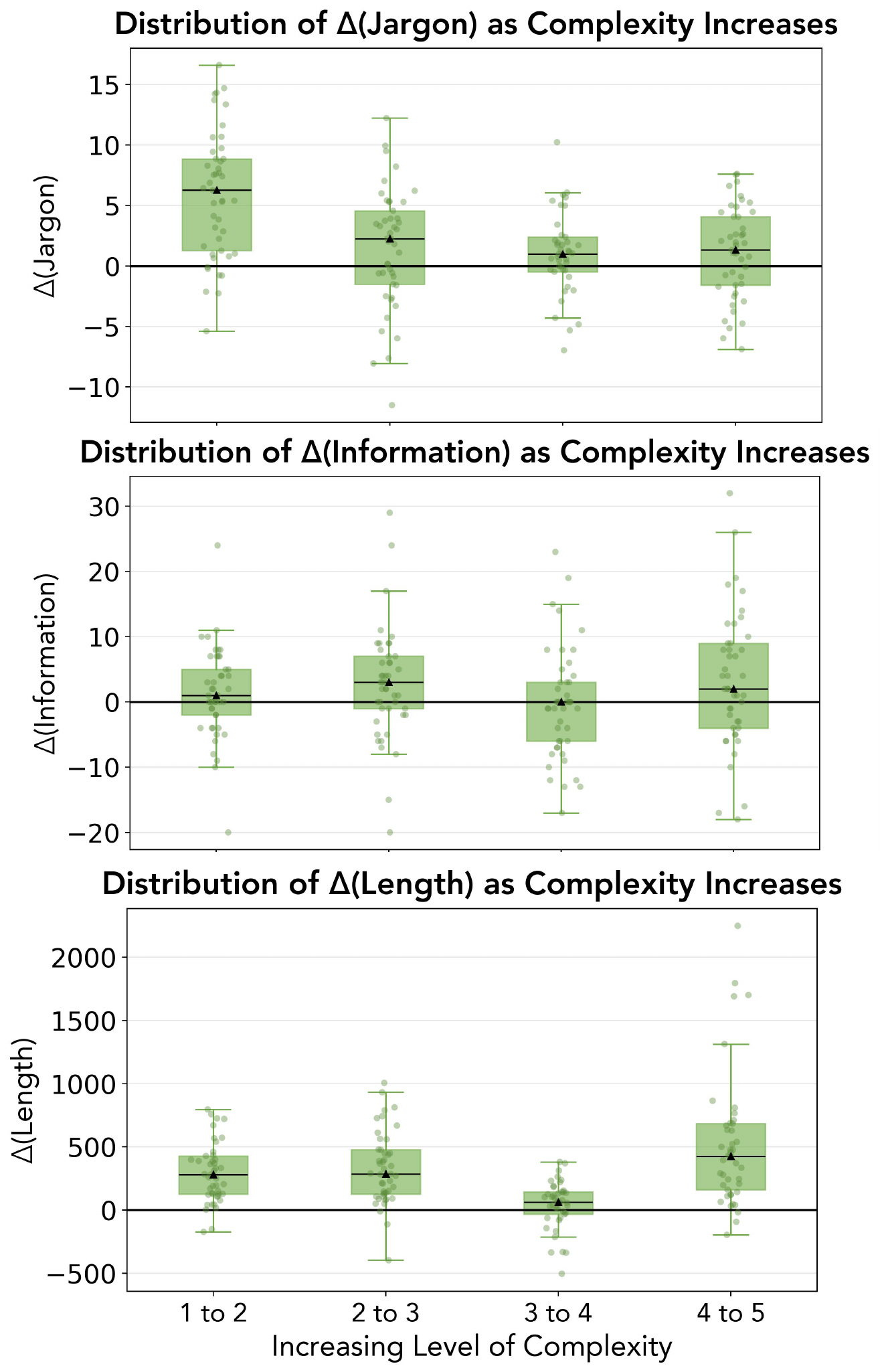}
    \caption{\textbf{Model performance during user study based on complexity measures} Between consecutive levels of complexity, the interactive condition had comparable changes in \textsc{Jargon}, \textsc{Information}, and \textsc{Length} to the model evaluation (Fig. \ref{fig:boxplots}), varying between increasing and decreasing. We use GPT-5 mini to supply the responses as a low-latency option. Each point in the scatter overlay represents one of the $45$ participant queries. Ideally, all measures should increase between levels (i.e., all points should be strictly above the zero line). Extreme outliers removed for visualization.}
    \label{fig:user_study_complexity}
\end{figure}

\subsubsection{User Study Interview Guide}
\label{sec:interview_guide}
We include the interview guide we used to structure questions during the user study.
The questions aim to probe at participants' perceptions of and interactions with both interfaces.

The following questions were asked for both conditions:
    \begin{itemize}[noitemsep,topsep=0pt]
        \item How did you feel about the level of complexity in the chat responses? Were there times that you felt that you had too much or too little information? Was the amount of complexity appropriate or overwhelming?
        \item How did you use the system generally? What worked well and what didn't?
        \item What about the text in the responses would you change if anything?
        \item How did you feel about not reading the papers? (if the participant expressed something about this)
    \end{itemize}
    
The following questions were asked only for the interactive condition:
    \begin{itemize}[noitemsep,topsep=0pt]
        \item How did you feel about having a choice of 5 responses with varying complexity as opposed to one response with fixed complexity? [after participants had experienced both conditions]
        \item When did you find the slider helpful or not helpful?
        \item When did you ask follow-up questions versus use the slider versus use the response as is?
        \item Is there anything about the progression of the text between the 5 levels that you would want to control?
        \item Did the variations in complexity match what you expected from the 5 levels? If not, what would you have wanted?
        \item Was there anything hard about going through the different levels?
        \item Did you read all 5 levels, why or why not?
        \item Could you tell the difference between the 5 levels [or the levels you did read] and how?
        \item Does 5 levels feel appropriate?
        \item Do you want the slider for every response?
    \end{itemize}

\subsubsection{Participant Details}
\label{sec:participants}
We recruited participants primarily through email listservs and Slack channels at multiple institutions.
Tab. \ref{tab:participants} lists participants' background and LLM usage.
Most participants held an academic affiliation and had a STEM background.
To ascertain general LLM usage, we asked ``How often do you use LLMs and LLM-infused applications?'', and the choices were:
\begin{itemize}[nosep]
    \item Never
    \item Rarely, about 1–2 times a month
    \item Sometimes, about 3–4 times a month
    \item Often, about twice a week
    \item Always, about once or more a day
\end{itemize}
Some wording came from \citet{kim2025reliance}.

For LLM usage in research, we asked ``Do you use LLMs and LLM-infused applications for any parts of the research process? Select all that apply.'', and the choices were:
\begin{itemize}[nosep]
    \item Yes, for information seeking (e.g., discovering papers, generating summaries, discovering topics)
    \item Yes, for editing writing (e.g., fixing grammar or rephrasing, looking up synonyms, formatting papers)
    \item Yes, for direct writing (e.g., rewriting to another style, shortening, summarizing)
    \item Yes, for data cleaning \& analysis (e.g., cleaning and reformatting data, statistical reporting, qualitative analysis)
    \item Yes, for ideation \& framing (e.g., brainstorming research questions, coming up with ways to frame a paper, getting inspiration for methods)
    \item Yes, for data generation (e.g., generating synthetic data, producing examples and labels)
    \item Yes, for other purposes (Select this box and Other and specify below in Other)
    \item No
    \item Other:
\end{itemize}
These answer choices mostly came from \citet{liao2025llms}.



\begin{table*}
\begingroup 
\renewcommand{\arraystretch}{1.5} 
\setlength{\tabcolsep}{10pt}    
    \centering
    \small
    \begin{tabular}{p{3cm}|p{3cm}|p{1.5cm}|p{1cm}|p{5cm}}
        \textbf{Background} & \textbf{Field of Study} & \textbf{Research} \newline \textbf{Experience} & \textbf{LLM \newline Usage} & \textbf{LLM Research Usage} \\
    \hline
         Engineer & Computer Science & 7 years & Always & Information seeking, Ideation \& framing, Other: Coding, Indexing \\

         Postdoctoral researcher & Comparative Literature & 8 years & Rarely & Information seeking, Ideation \& framing \\
         
         1st year Ph.D. student & Computer Science & 1 year & Rarely & None \\
         
         Master's student & Computer Science & 2 years & Always & Information seeking, Editing writing, Direct writing, Data cleaning \& analysis, Ideation \& framing \\
    
         2nd year Ph.D. student & Computer Science & 4 years & Always & Information seeking, Editing writing, Direct writing, Data cleaning \& analysis, Ideation \& framing, Data generation, Other: Getting feedback on paper drafts \\
    
         Undergraduate student & Statistics, Actuarial Science & None & Often & Information seeking, Editing writing, Direct writing, Data cleaning \& analysis, Ideation \& framing \\
    
         5th year Ph.D. student & Applied Mathematics & 4 years & Always & Information seeking, Editing writing, Direct writing \\
    
         Master's student & Computer Science & 2 years & Always & Information seeking, Editing writing \\
    
         Master's student & Computer Science & 1 year & Always & Information seeking, Editing writing, Ideation \& framing \\
    
         4th year Ph.D. student & Computer Science & 5 years & Often & Information seeking, Other: Coding \\
    
         4th year Ph.D. student & Computer Science & 4 years & Always & Information seeking, Editing writing, Direct writing, Data cleaning \& analysis, Ideation \& framing \\
    
         2nd year Ph.D. student & Computational Biology & 4 years & Often & Information seeking, Data cleaning \& analysis \\
    
         2nd year Ph.D. student & Computer Science & 2 years & Always & Editing writing, Other: Coding \\
    
         College Graduate & Statistics, Computer Science & None & Always & Information seeking, Editing writing, Data cleaning \& analysis \\
    
         Master's student & Electrical \& Computer Engineering & 2 years & Always & Information seeking, Editing writing, Ideation \& framing \\
    
         5th year Ph.D. student & Computer Science & 6 years & Always & Editing writing, Ideation \& framing \\
    \bottomrule
    \end{tabular}
    \caption{\textbf{Participants' background and LLM usage}}
    \label{tab:participants}
    \endgroup
\end{table*}

\subsubsection{Qualitative Coding Procedure}
\label{sec:qual_coding_app}
As mentioned in Sec. \ref{sec:userstudy_method}, to analyze the qualitative data from the study transcripts, we employ open coding \citep{Saldana2021coding}.
We follow work where multiple authors create a codebook from sampled interviews and one author codes the remaining interviews with the codebook \cite{kim2023hmhta,McDonald2019irr}.
Specifically, one author created an initial codebook from $4$ studies randomly sampled evenly between conditions, which all authors then iterated on until a final codebook was agreed upon (Tab. \ref{tab:codebook}).
Using the final codebook, the same initial author coded the remaining studies.

\begin{table*}[bt]
    \centering
    \small
    \begin{tabular}{p{4cm}|p{3cm}|p{7cm}}
    \toprule
         \textbf{Theme} & \textbf{Sample Code} & \textbf{Sample Quote} \\
         \midrule
             Complexity is situational & Complexity is situational & P3: ``\textit{I liked the option to see different complexities for a topic because again sometimes I might want more complex, sometimes I might want less complex depending on the task, depending on what I'm actually trying to learn.}'' \\ 
        \midrule
             Inconsistent satisfaction with the conventional chatbot & Conventional: inconsistent satisfaction with answer & P13: ``\textit{sometimes you're getting the right complexity with some sentences and sometimes you're not. Yeah...it's not super consistent in its complexity.}'' \\
        \midrule
             Users appreciate the convenience and agency that interactive complexity adds & Appreciates agency of interactive chat & P9: ``\textit{I like the amount of choice. I feel like if I don't use it, I don't use it, but I like having it there if I want to use it.}'' \\
        \midrule
             Users look for decreasing jargon, information, and length as they decrease the level of complexity & Want less information when decreasing complexity & P9: ``\textit{I want it to be shorter. I think shorter and honestly like less information.}'' \\
        \midrule
            While 5 levels of complexity seemed generally appropriate, the levels need to be more distinct from each other & Levels are too similar & P15: ``\textit{I didn't notice like a meaningful enough jump between each level}'' \\
         \bottomrule
    \end{tabular}
    \caption{\textbf{Codebook for analyzing our formative study} Shown is a list of themes with sample codes and quotes. Some codes match the themes as they formed themes on their own.}
    \label{tab:codebook}
\end{table*}

\subsection{Validity checks}
\label{sec:validity_app}
\subsubsection{Manual validation of Information measure}
\label{sec:info_validation_app}
We used the prompt provided by \citet{min2023factscore} for identifying what they call ``atomic facts'' (defined as ``pieces of information''). 
Since we used a more recent model (GPT-4.1 instead of InstructGPT), we manually validated a randomly-sampled subset of $25$ responses according to the following procedure and achieved $80\%$ agreement between the Information measure and the human rater:
\begin{enumerate}
    \item One author manually identified all atomic facts following the procedure and definition from \citet{min2023factscore}.
    \item The number of manually-identified facts was compared to the number of atomic facts that GPT-4.1 identified.
    \item If the GPT-4.1 number was $\pm 2$ of the author’s number, it was considered consistent.
\end{enumerate}

\subsubsection{Factuality}
\label{sec:factuality_app}
We run an evaluation of factual correctness using SummaC \cite{Laban2022SummaC} to compare the model responses to the human-written reports that were used. SummaC is a popular NLI-based measure of how much a summary follows from a source; higher scores indicate that more of the information in the original report is retained and correct. We find that the responses have moderate scores, consistent across levels within the same model (Tab. \ref{tab:factuality}). This suggests, unsurprisingly, that information is lost in text simplification and to a similar degree across levels.
\renewcommand{\arraystretch}{1}
\begin{table*}[h]
    \centering
    \small
    \begin{tabular}{l|c|c|c|c|c}
        \textbf{Model} & \textbf{Level 1} & \textbf{Level 2} & \textbf{Level 3} & \textbf{Level 4} & \textbf{Level 5} \\
        \midrule
             GPT-5.1 & 0.46 ± 0.13 & 0.45 ± 0.12 & 0.46 ± 0.13 & 0.46 ± 0.15 & 0.45 ± 0.14  \\
             GPT-5 mini & 0.44 ± 0.13 & 0.41 ± 0.11 & 0.4 ± 0.11 & 0.39 ± 0.11 & 0.39 ± 0.12 \\
             Claude Sonnet 4.5 & 0.45 ± 0.12 & 0.45 ± 0.13 & 0.44 ± 0.14 & 0.41 ± 0.13 & 0.39 ± 0.13 \\
             \makecell{ + Thinking} & 0.46 ± 0.11 & 0.47 ± 0.13 & 0.42 ± 0.12 & 0.41 ± 0.12 & 0.4 ± 0.13 \\
             DeepSeek-V3.1 & 0.46 ± 0.12 & 0.44 ± 0.12 & 0.41 ± 0.14 & 0.4 ± 0.11 & 0.38 ± 0.11 \\
         \midrule
    \end{tabular}
    \caption{\textbf{Average factuality scores across models and levels of complexity.} Means and standard deviations of distributions of SummaC scores are shown (using ``College $\rightarrow$ Sr. Res., Expert-written Inputs (n=98)'' evaluation).}
    \label{tab:factuality}
\end{table*}

\subsubsection{Consistency between runs}
To check if the performance across multiple model calls is consistent, we run the College → Sr. Res. (n=98) evaluation 5 times using Claude Sonnet 4.5 (one of our better-performing models).
We find that the performance is consistent across the 5 runs (Tab. \ref{tab:consistency}).
\renewcommand{\arraystretch}{1}
\begin{table*}[h]
    \centering
    \small
    \begin{tabular}{l|c|c|c|c}
        \textbf{Measure} & \textbf{1 to 2} & \textbf{2 to 3} & \textbf{3 to 4} & \textbf{4 to 5} \\
        \midrule
             Jargon & 100 ± 0 & 82.68 ± 2.56 & 66.80 ± 3.88 & 63.09 ± 5.88  \\
             Information & 80.41 ± 2.44 & 81.24 ± 2.87 & 81.03 ± 2.66 & 82.68 ± 1.52 \\
             Length & 100 ± 0 & 100 ± 0 & 100 ± 0 & 100 ± 0 \\
         \midrule
    \end{tabular}
    \caption{\textbf{Consistency in performance between separate model calls.} Means and standard deviations of the distribution of per-transition model performances, similar to Tab. \ref{tab:model_performance_percentages}, across 5 calls to Claude Sonnet 4.5 are shown (using ``College $\rightarrow$ Sr. Res., Expert-written Inputs (n=98)'' evaluation).}
    \label{tab:consistency}
\end{table*}

\end{document}